\documentclass{article}

\PassOptionsToPackage{numbers,compress}{natbib}

\usepackage[preprint]{nips_requirements/neurips_2026}

\makeatletter
\renewcommand{\@notice}{}
\makeatother

\usepackage[table]{xcolor}
\usepackage[utf8]{inputenc}
\usepackage[T1]{fontenc}
\usepackage{url}
\usepackage{booktabs}
\usepackage{amsfonts}
\usepackage{amsmath}
\usepackage{amssymb}
\usepackage{nicefrac}
\usepackage{microtype}
\usepackage[table]{xcolor}
\usepackage{graphicx}
\usepackage{wrapfig}
\usepackage{multirow}
\usepackage{makecell}
\usepackage{threeparttable}
\usepackage{array}
\usepackage{subcaption}
\usepackage{adjustbox}
\usepackage{tabularx}
\usepackage{pdflscape}
\usepackage{enumitem}
\usepackage{float}

\graphicspath{{./}{./figs/}}

\usepackage{hyperref}

\newcommand{\best}[1]{\textbf{#1}}
\newcommand{\second}[1]{\underline{#1}}
\newcommand{\ours}[1]{\textbf{#1}}

\title{GlobalForge: Towards Robust AI-Generated Image Detection}
\date{}

\author{
  \normalfont\normalsize
  Manni Cui\textsuperscript{1,*} \quad
  Ruiqi Liu\textsuperscript{2,*} \quad
  Dianyuan Zou\textsuperscript{1,*} \quad
  Ziheng Qin\textsuperscript{2} \quad
  Jingrui Xu\textsuperscript{1} \quad
  ZiAn Wang\textsuperscript{3} \\
  Jianglan Wei\textsuperscript{1} \quad
  Han Zhou\textsuperscript{1} \quad
  Yu Liu\textsuperscript{1} \quad
  Yan Wang\textsuperscript{4,\textdagger} \quad
  Shu Wu\textsuperscript{2,\textdagger} \\[12pt]
  \makebox[\dimexpr\textwidth-2\tabcolsep\relax][c]{%
    \textsuperscript{1}Huazhong University of Science and Technology \quad
    \textsuperscript{2}Institute of Automation, Chinese Academy of Sciences} \\
  \makebox[\dimexpr\textwidth-2\tabcolsep\relax][c]{%
    \textsuperscript{3}Jilin University \quad
    \textsuperscript{4}Tsinghua University} \\[3pt]
  \textsuperscript{*}Equal contribution \quad
  \textsuperscript{\textdagger}Corresponding author
}

\begin{document}

\maketitle
\begin{figure}[H]
\centering
\includegraphics[width=\linewidth]{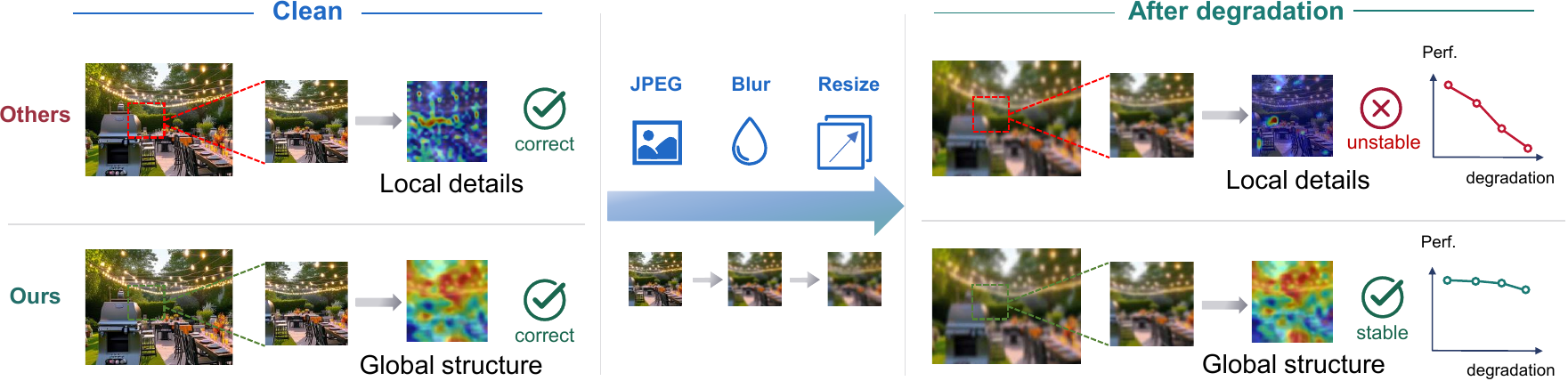}
\vspace{10pt}
\caption{Under clean conditions, existing detectors and GlobalForge both correctly identify AI-generated images. After common real-world degradations (JPEG compression, blur, resize), other methods fail because their local artifact cues are destroyed; GlobalForge maintains correct detection by relying on degradation-stable global structural features.}
\label{fig:teaser}
\end{figure}
\vspace{10pt}
\begin{abstract}
AI-generated image (AIGI) detectors achieve strong accuracy on clean benchmarks, but their performance drops sharply after images are propagated through real-world channels. We trace this fragility to what these detectors actually learn: they overfit to local artifacts left by generators in small spatial neighborhoods, which are easily destroyed by common propagation degradations such as JPEG compression and blur. Instead, we shift the discriminative cue from fragile local artifacts to more robust global structure. Building on this, we propose GlobalForge, a framework with two complementary modules. The Local Information Bottleneck (LIB) suppresses local components to block shortcut learning, while the Global Structural Reasoning (GSR) module forces every token to gather evidence from distant regions. Both modules are trained jointly under a contrastive structural loss based on degradation that keeps the resulting features stable under degradation. To support fine-grained robustness evaluation, we further introduce RealDeg-Bench, covering 7 common degradation operators and multi-step compound chains. GlobalForge improves average BAcc on 8 in-the-wild benchmark groups by $\mathbf{5.89\%}$ over the previous state-of-the-art, and is clearly ahead of representative baselines on RealDeg-Bench under both single and compound degradations. Code is available at \url{https://anonymous.4open.science/r/GlobalForge-BE0F/}.
\end{abstract}

\vspace*{\fill}
\newpage
\section{Introduction}
\label{sec:intro}

Recent generative models can synthesize images that are visually indistinguishable from real photographs. This has raised serious concerns about deepfakes, disinformation, and copyright infringement~\cite{mirsky2021creation,tolosana2020deepfakes}, and reliable detection of AI-generated images (AIGIs) has become an urgent need. Many detection methods have been proposed, and most achieve strong accuracy on clean benchmarks. But real deployment is far harsher. Images shared on social media are repeatedly compressed, color-adjusted and re-edited. These operations build up into compound degradation chains~\cite{corvi2023detection,cozzolino2024clip}, making current detectors lose substantial accuracy~\cite{pal2024semi}.

The drops are not random failures. They reveal that the cue these detectors learn from is the wrong one. Such detectors overfit to local artifacts left by generators in small spatial neighborhoods, and any propagation degradation can destroy these artifacts directly. We give a detailed diagnosis in Section~\ref{sec:motivation}. Most prior work still treats local artifacts as the right signal to defend. Data augmentation~\cite{yan2024sanity,pal2024semi} broadens the training distribution but remains a passive defense, and prior work~\cite{pal2024semi} shows it generalizes poorly to unseen degradations. Image restoration~\cite{ke2023df} tries to recover damaged artifacts, but they already lie outside the natural-image distribution and restoration pulls the image back toward natural-image statistics. Neither line addresses the problem at its root.

We argue for a different path. Instead of defending fragile local artifacts, we intervene in what the model is allowed to learn from, and steer it toward global structural coherence, by which we mean cues that depend on long-range relationships across the image rather than on any single local neighborhood. Such cues are not easy to disrupt with local perturbations, and they do not depend on the local signature of any specific generator, so they help with degradation robustness and cross-generator generalization at the same time.

Following this idea, we propose GlobalForge, a detection framework with two complementary modules. A Local Information Bottleneck (LIB) suppresses the fragile local components on which shortcut learning would otherwise latch. A Global Structural Reasoning (GSR) module forces every token to gather evidence from distant regions, so the model has to rely on long-range structure for the decision. The two modules are trained jointly under a degradation-aware contrastive structural loss ($\mathcal{L}_{\text{DCS}}$) that further keeps the resulting global representation stable under realistic degradation. Our contributions are summarized as:
\vspace{-2pt}
\begin{itemize}[leftmargin=1.2em,itemsep=1pt,topsep=1pt,parsep=0pt,partopsep=0pt]
    \item We identify local-artifact dependency as the root cause of degradation fragility in existing AIGI detectors, and a controlled experiment shows that blocking the model's access to local artifacts improves both degradation robustness and cross-generator generalization.
    \item We propose \textbf{GlobalForge}, to our knowledge, the first AIGI detection framework that actively suppresses local shortcuts and forces the discriminative decision onto long-range structural cues, achieving a robustness profile that is decoupled from generator-specific local signatures.
    \item We introduce \textbf{RealDeg-Bench}, a robustness-oriented evaluation benchmark for fine-grained diagnosis of AIGI detectors under common single-operator degradations and multi-step compound chains that approximate real social-media propagation.
    \item GlobalForge improves average BAcc on 8 in-the-wild benchmark groups by $\mathbf{5.89\%}$ over the previous state-of-the-art, and reaches $\mathbf{85.81\%}$ average BAcc on RealDeg-Bench, clearly ahead of representative baselines under both single and compound degradations.
\end{itemize}

\begin{figure*}[t]
\centering

\begin{minipage}[c]{0.47\textwidth}
  \centering
  \includegraphics[width=\linewidth]{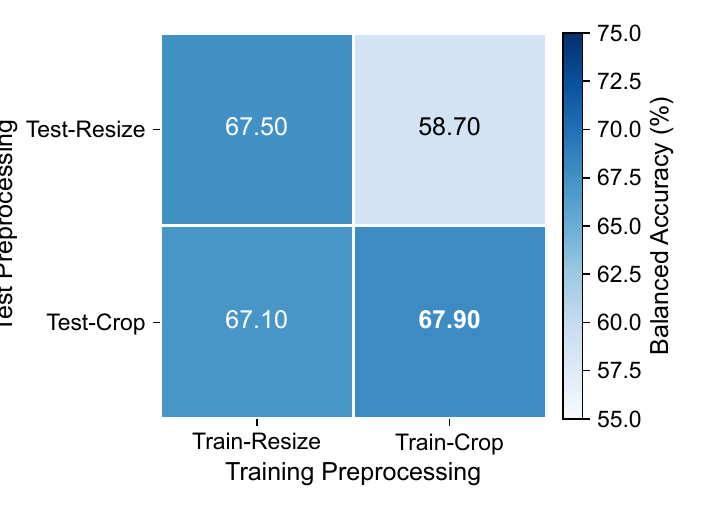}\end{minipage}
\hfill
\begin{minipage}[c]{0.47\textwidth}
  \centering
  \vspace*{-40pt}   \includegraphics[width=\linewidth]{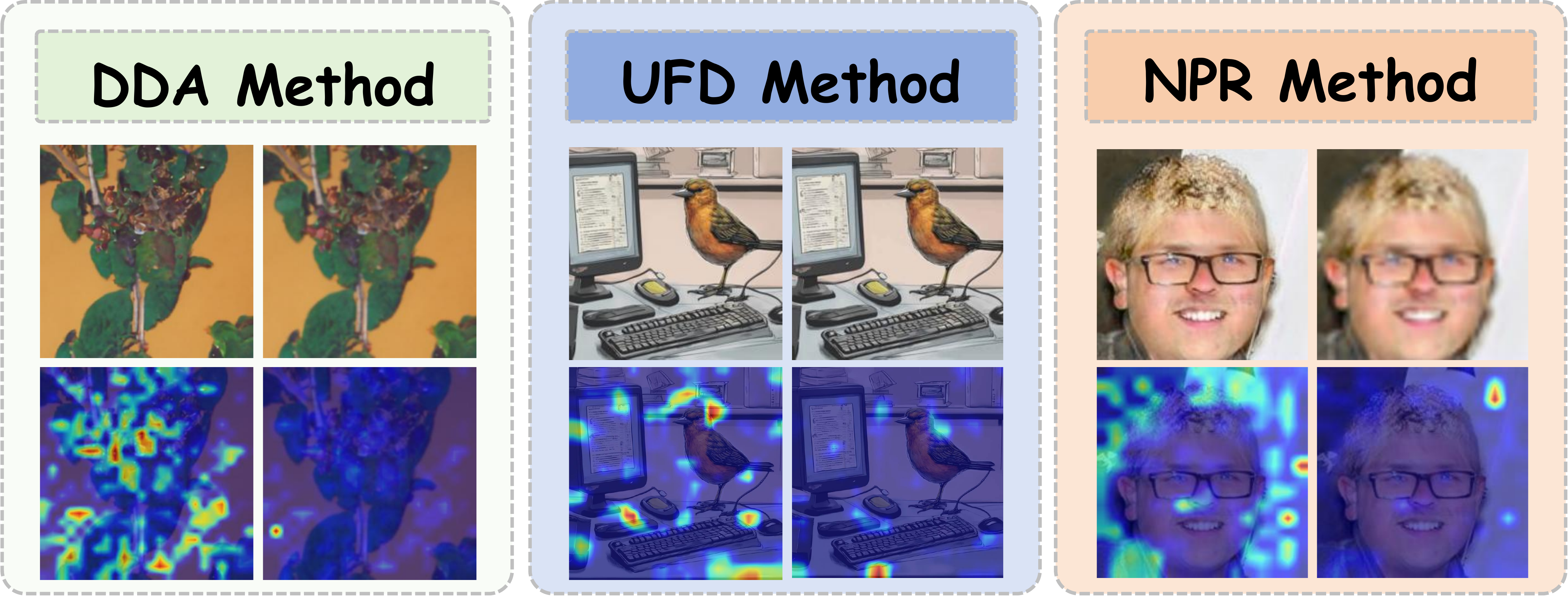}\end{minipage}

\par\vspace*{-45pt}\noindent

\begin{minipage}[b]{0.47\textwidth}
  \captionof{figure}{
    \textbf{Preprocessing cross-experiment (UnivFD on GenImage).}
    Rows: test-time preprocessing; columns: training-time.
    Crop-trained models suffer a sharp accuracy drop when evaluated with resize,
    revealing reliance on spatially sensitive local cues.
  }
  \label{fig:crop-resize}
\end{minipage}
\hfill
\begin{minipage}[b]{0.47\textwidth}
  \captionof{figure}{
    \textbf{Attention collapse under image degradation.}
    Grad-CAM visualizations for DDA, UnivFD, and NPR under different degradations.
    For each method, the top row shows the inputs and the bottom row shows the
    corresponding attention maps; within each method, the left column is the clean
    image and the right column is its degraded counterpart.
    The three methods are evaluated under resizing, JPEG compression, and blur,
    respectively.
  }
  \label{fig:attention-collapse}
\end{minipage}

\end{figure*}

\section{Related Work}
\label{sec:related}

\subsection{AI-Generated Image Detection}

Existing AIGI detectors fall into three paradigms. \textbf{Pretrained-representation-based} detectors freeze or lightly adapt the features of large pretrained models such as CLIP, and treat real/fake detection as a binary classification task on top of these general-purpose representations~\cite{wang2020cnn,ojha2023towards,liu2024forgery,tan2025c2p,yan2024sanity,chen2025dual,qin2025scaling}. \textbf{Pixel or frequency statistics based} detectors target low-level generative fingerprints such as spectral artifacts~\cite{frank2020leveraging}, upsampling grid traces~\cite{tan2024rethinking}, frequency inconsistency~\cite{li2025improving}, and diffusion reconstruction error~\cite{wang2023dire,chen2024drct}. \textbf{Semantic and structure based} methods such as LASTED~\cite{wu2025generalizable} and AntifakePrompt~\cite{chang2023antifakeprompt} instead use higher-level signals, and share the global-consistency spirit advocated in this work.

A growing line of work directly targets robustness under propagation degradations. AIDE~\cite{yan2024sanity} and Semi-Truths~\cite{pal2024semi} broaden training distributions through aggressive degradation augmentation, but Semi-Truths itself reports that augmentation alone does not shift detectors' feature preference. B-Free~\cite{guillaro2025bias} addresses platform-induced bias through a debiased training paradigm, while DF-UDetector~\cite{ke2023df} restores degraded inputs to recover artifacts, yet the restored statistics drift toward natural images rather than the abnormal traces of generators. TextureCrop~\cite{konstantinidou2025texturecrop} and Karageorgiou et al.~\cite{karageorgiou2025any} go the opposite way and reinforce the local-artifact cue, which helps on clean benchmarks but widens the gap once the image is degraded. But these efforts remain largely passive. They treat robustness as a downstream issue to be patched through input-side or training-side compensation, and rarely examine why the underlying discriminative cue is fragile in the first place. As a result, the model's feature preference is left untouched, and accuracy still degrades sharply along realistic propagation chains.

\subsection{AIGI Detection Evaluation Benchmarks}

Early AIGI benchmarks mostly target cross-generator generalization. ForenSynths~\cite{wang2020cnn} evaluates detectors on GANs, and GenImage~\cite{zhu2023genimage} extends this to million-scale diffusion images across multiple generators, becoming the de facto training and evaluation suite for modern detectors. Chameleon~\cite{yan2024sanity} further tracks detector performance against rapidly evolving generators. More recent efforts move toward deployment-oriented evaluation. WildRF~\cite{cavia2024real} and BFree-Online~\cite{guillaro2025bias} collect images directly from social-media platforms to approximate in-the-wild conditions, while real-chain~\cite{liu2025beyond} captures images that have already traversed multi-step social-platform degradations.

These benchmarks have two structural limitations for studying degradation robustness. First, their robustness protocols typically apply only one perturbation at a time, such as JPEG at a fixed quality factor or Gaussian blur at a single strength~\cite{corvi2023detection,cozzolino2024clip}, which makes it hard to characterize how accuracy decays along realistic degradation chains. Second, in-the-wild benchmarks rely on degradations that arise incidentally from platform processing, with uncontrollable type, strength, and depth, so they cannot pinpoint which operators or compositions drive the failures.

\section{Motivation and Design Principles}
\label{sec:motivation}

As discussed in Section~\ref{sec:intro}, AIGI detectors that perform near-saturation on clean benchmarks routinely collapse once images traverse real propagation chains. We argue that this fragility is a structural property of what existing detectors learn. Generators leave behind local artifacts in small spatial neighborhoods~\cite{frank2020leveraging,corvi2023intriguing,tan2024rethinking}, and these artifacts all sit in the local part of the image. Common propagation degradations such as JPEG quantization, resize resampling, and blur smoothing all perturb the local spectrum. They therefore destroy these artifacts with little effort. The artifacts are also generator-specific, since different generators apply different local operators, so detectors trained on them struggle on unseen generators. They behave less like useful cues and more like shortcuts~\cite{geirhos2018imagenet,geirhos2020shortcut} that crowd out more reliable evidence.

Common practice in the field already points in this direction. Existing AIGI detectors almost universally adopt center crop rather than resize at inference time~\cite{wang2020cnn,ojha2023towards,tan2024rethinking}, because resize causes a substantial drop in detection accuracy. Yet resize preserves both the semantic content and the spatial layout of an image, so the cue it disrupts cannot be semantic or structural in nature. A preprocessing cross-experiment that varies training and test time preprocessing independently confirms this asymmetry directly (Fig.~\ref{fig:crop-resize}). A model trained with center crop loses about \textbf{8.8\%} of BAcc when tested with resize, while a model trained with resize barely degrades when tested with resize at all. The cue that collapses must therefore be a fine-grained signal sensitive to the spatial arrangement of pixel neighborhoods, in agreement with the local-artifact characterization above.

The same fragility extends from resize to other common degradations and shows up directly in the model's attention. Grad-CAM visualizations of three representative detectors (Fig.~\ref{fig:attention-collapse}) reveal that on clean images, the attention of all three concentrates sharply on local patches. Once a mild degradation is applied, these focal regions dissolve into diffuse, noise-like maps. The discriminative basis these detectors rely on vanishes together with the local artifacts, exactly as our diagnosis predicts.

This diagnosis motivates GlobalForge's two core design principles: suppress local information through LIB and explicitly guide global structural reasoning through GSR.

\section{Method}
\label{sec:method}
 
\begin{figure}[t]
\centering
\includegraphics[width=\linewidth]{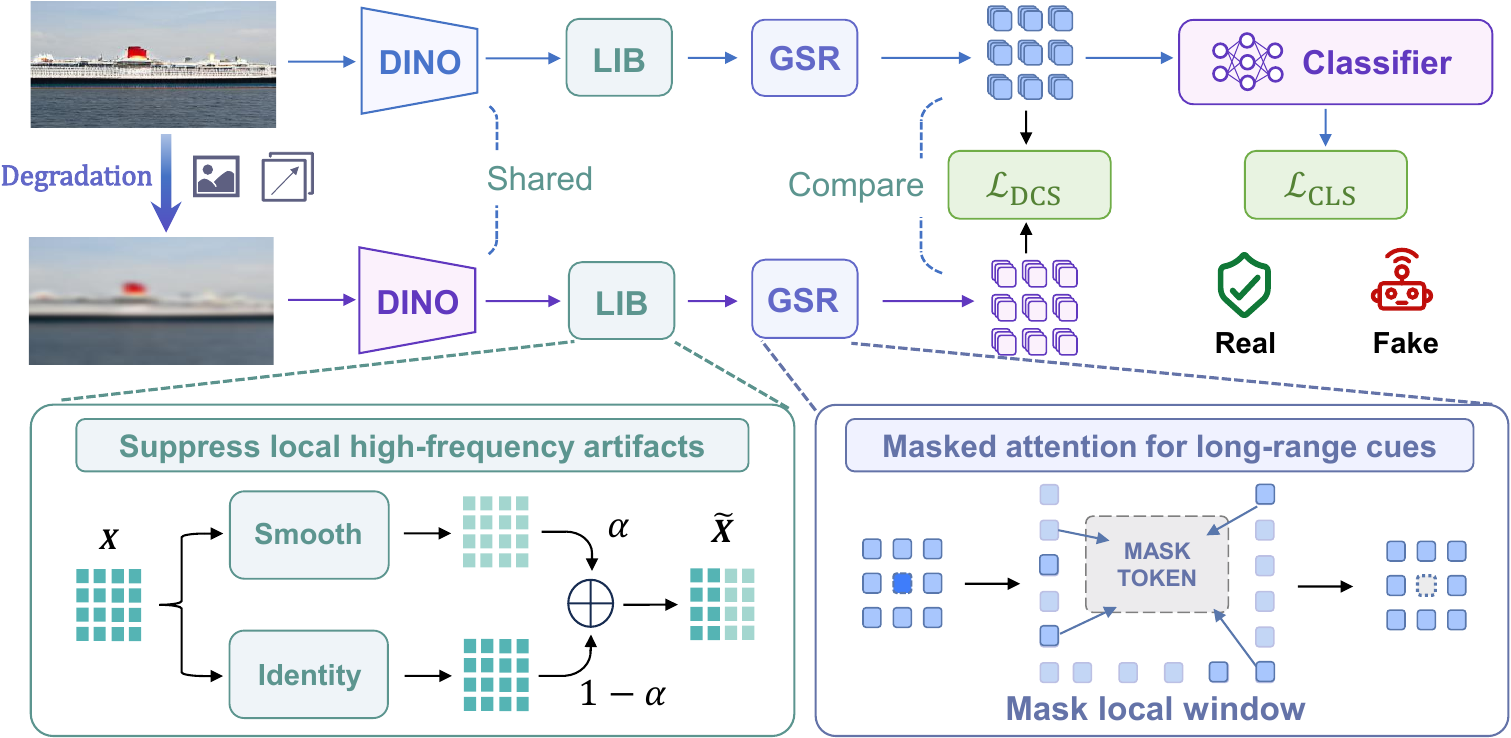}
\caption{Overall architecture of GlobalForge.}
\label{fig:framework}
\end{figure}

The overall architecture of GlobalForge is shown in Figure~\ref{fig:framework}. Given an input image $I \in \mathbb{R}^{3 \times H_0 \times W_0}$, our goal is to learn a discriminative mapping $f_\theta: I \to y \in \{0, 1\}$ that remains reliable under real-world degradation. The backbone features first pass through \textbf{LIB} and then \textbf{GSR}, and are read out by a classification head. The two modules are trained jointly under a degradation-aware contrastive structural loss ($\mathcal{L}_{\text{DCS}}$) that aligns clean and degraded views in feature space. The components are detailed in Sections~\ref{sec:method-lib}--\ref{sec:method-optim}.

\subsection{Local Information Bottleneck (LIB)}
\label{sec:method-lib}

To prevent the model from overfitting to fragile local artifacts, LIB introduces a learnable ``original-feature / smoothed-feature'' fusion pathway on the deep features of the backbone, thereby suppressing local high-frequency components in a controlled manner.

A natural alternative is to apply low-pass filtering in the pixel or frequency domain of the input image. We instead operate in the feature domain. Further comparison of these input-side alternatives with LIB is reported in Section~\ref{sec:ablation}, where pixel-domain blur and input-frequency low-pass filtering both underperform the feature-domain bottleneck. Filtering directly in the spatial or frequency domain of the image leaves visible side effects, such as ringing and Gibbs-like oscillations near edges, which the model can latch onto as new shortcuts and which themselves shift under degradation. Operating on deep features avoids this: the suppression is applied after the backbone has already encoded the image, so the bottleneck shapes \emph{what the classifier sees}, not the raw input distribution.

Specifically, let the patch tokens produced by the backbone be $\mathbf{F} \in \mathbb{R}^{B \times N \times d}$, and reshape them into a spatial feature map $\mathbf{X} \in \mathbb{R}^{B \times d \times h \times w}$, where $B$ is the batch size, $N = h \times w$ is the number of patch tokens, $d$ is the channel dimension, and $h, w$ are the spatial height and width of the feature map. We construct a fixed $k \times k$ Gaussian kernel (with $\sigma = k/3$) and obtain a smoothed version $\mathbf{X}_{\text{smooth}}$ via depthwise convolution. We introduce a single learnable scalar $\beta$ and map it through a sigmoid to a fusion coefficient $\alpha = \sigma(\beta) \in (0, 1)$. The final output is the convex combination of the two:
\begin{equation}
\hat{\mathbf{X}} = (1 - \alpha)\mathbf{X} + \alpha \mathbf{X}_{\text{smooth}}.
\end{equation}

Gaussian smoothing acts as a low-pass filter, attenuating the high-frequency feature components where local artifacts concentrate (upsampling grids, noise residues) while preserving the lower-frequency structural information. As $\alpha \to 0$ the output degenerates to the original feature; the underlying raw gate scalar $\beta$ (from which $\alpha = \sigma(\beta)$ is derived) is optimized jointly with the rest of the model.

\subsection{Global Structural Reasoning (GSR)}
\label{sec:method-gsr}

LIB suppresses local high-frequency components along the frequency axis, but the model may still exploit local correlations between spatially adjacent tokens. GSR closes this loophole by forcing each token to gather evidence exclusively from distant regions, complementing LIB along the spatial axis.

Specifically, the LIB output $\hat{\mathbf{X}} \in \mathbb{R}^{B \times d \times h \times w}$ is reshaped back to a token sequence $\hat{\mathbf{F}} \in \mathbb{R}^{B \times N \times d}$. GSR first computes the standard self-attention score matrix:
\begin{equation}
\mathbf{S} = \frac{(\hat{\mathbf{F}} \mathbf{W}_q)(\hat{\mathbf{F}} \mathbf{W}_k)^\top}{\sqrt{d}}.
\end{equation}

To block the model from exploiting neighborhood local correlations, we apply a local masking window at each position before the Softmax: all attention entries whose Chebyshev distance (i.e., chessboard distance $d_{\text{Cheb}}(u, v) = \max(|u_x - v_x|, |u_y - v_y|)$) to the query position is at most $w_{\text{gsr}}$ (default $w_{\text{gsr}} = 3$) are set to $-\infty$:
\begin{equation}
\tilde{\mathbf{S}}_{u,v} = \begin{cases} -\infty, & \text{if } d_{\text{Cheb}}(u, v) \leq w_{\text{gsr}}, \\ \mathbf{S}_{u,v}, & \text{otherwise}. \end{cases}
\end{equation}

The masked scores are then normalized by Softmax, used to update the features, and passed through an output projection with a residual connection:
\begin{equation}
\mathbf{F}_{\text{GSR}} = \text{Proj}\big(\text{Softmax}(\tilde{\mathbf{S}}) \cdot \hat{\mathbf{F}} \mathbf{W}_v\big) + \hat{\mathbf{F}}.
\end{equation}

During training, masking is applied with probability $p_{\text{mask}}=1.0$ by default, with sensitivity to the masking probability and window size analyzed in Section~\ref{sec:sensitivity}. This design forces each token's representation update to come exclusively from distant regions, so the model can only discriminate by integrating cross-region structural information. Combined with the frequency-level suppression of LIB, this steers the model toward a global structural representation that is more stable under both degradation and generator shifts.

\subsection{Synergistic Optimization Objectives}
\label{sec:method-optim}

The overall training loss of GlobalForge consists of a classification term and a degradation-aware contrastive structural term:
\begin{equation}
\mathcal{L} = \mathcal{L}_{\text{cls}} + \lambda_{\text{dcs}} \mathcal{L}_{\text{DCS}},
\end{equation}
where $\mathcal{L}_{\text{cls}}$ is the binary cross-entropy loss with label smoothing, $\mathcal{L}_{\text{DCS}}$ is an auxiliary contrastive regularizer (defined below), and $\lambda_{\text{dcs}} = 0.01$ by default. At the optimizer level, all trainable parameters are managed by a single AdamW optimizer and partitioned into two parameter groups by weight-decay policy: a \emph{decay} group (projection matrices and linear layers) with weight decay $0.01$, and a \emph{no-decay} group (biases, layer-norm scales, and the LIB gate scalar $\beta$) with weight decay $0$.
$\mathcal{L}_{\text{DCS}}$ serves as an auxiliary regularizer that explicitly aligns paired clean and degraded views at the representation level. For each input $I_i$, we online-generate a compound-degraded view $\tilde{I}_i$ (random JPEG compression, Gaussian blur, and color perturbation), and denote the global image representations of the clean and degraded views as $\mathbf{z}_i$ and $\tilde{\mathbf{z}}_i$. $\mathcal{L}_{\text{DCS}}$ then takes the symmetric InfoNCE~\cite{oord2018representation} form:
\begin{equation}
\mathcal{L}_{\text{DCS}} = -\frac{1}{2B} \sum_{i=1}^{B} \left[ \log \frac{\exp(\mathrm{sim}(\mathbf{z}_i, \tilde{\mathbf{z}}_i) / \kappa)}{\sum_j \exp(\mathrm{sim}(\mathbf{z}_i, \tilde{\mathbf{z}}_j) / \kappa)} + \log \frac{\exp(\mathrm{sim}(\tilde{\mathbf{z}}_i, \mathbf{z}_i) / \kappa)}{\sum_j \exp(\mathrm{sim}(\tilde{\mathbf{z}}_i, \mathbf{z}_j) / \kappa)} \right],
\end{equation}
where $\mathrm{sim}(\cdot,\cdot)$ is cosine similarity and $\kappa$ is a temperature coefficient. This objective encourages the model to preserve degradation-invariant structural information rather than letting degradation signals leak into the final embedding.

\section{RealDeg-Bench: A Compound-Degradation Robustness Benchmark}
\label{sec:realdeg-bench}

\paragraph{Motivation.}
Existing AIGI robustness benchmarks either test one perturbation at a time or rely on in-the-wild images with uncontrolled degradation types, strengths, and histories. However, real image propagation often involves compound degradation chains, such as re-compression, resizing, and re-uploading. To better characterize robustness under such realistic propagation processes, we introduce \textbf{RealDeg-Bench}, a compound-degradation benchmark that evaluates detector performance across controllable degradation depths and strengths.

\paragraph{Construction.}
RealDeg-Bench is built on T2I-CoReBench~\citep{li2025easier}, which covers mainstream generators across GANs, diffusion models, and visual autoregressive architectures. We define a pool of seven degradation operators commonly observed in real propagation: JPEG compression, Gaussian blur, resize, Gaussian noise, brightness, contrast, and saturation, to cover real-world propagation conditions (full lists in Appendix~\ref{app:realdeg-bench}).

\paragraph{Evaluation protocol.}
RealDeg-Bench provides two complementary regimes. \emph{Single-operator} conditions apply one operator per image, while strength-sweep curves at each fixed level are also reported (Figure~\ref{fig:single-deg-curves}). \emph{Compound} conditions chain $N \in \{1,\dots,5\}$ operators sampled independently and uniformly from the seven-operator pool \emph{with replacement}, with each operator's strength independently drawn from its strength set:
\begin{equation}
\tilde{I} = \mathcal{D}_{k_n}\!\left(\cdots \mathcal{D}_{k_1}(I;\theta_1)\cdots;\theta_n\right), \quad k_i \sim \text{Uniform}(\mathcal{P}),\ \theta_i \sim \text{Uniform}(\mathcal{S}_{k_i}),
\end{equation}
where $\mathcal{P}$ is the operator pool and $\mathcal{S}_{k_i}$ is the strength set of operator $k_i$. With-replacement sampling allows the same operator to appear multiple times in a chain at different strengths, reflecting realistic propagation where an image is repeatedly re-compressed, re-resized, or re-color-shifted as it traverses successive platforms. In total, RealDeg-Bench contains 13 conditions (1 clean, 7 single-operator, 5 compound) with 95{,}589 images. Construction details, exact strength sets, and dataset statistics are provided in Appendix~\ref{app:realdeg-bench}.

  \begin{table}[t]
  \centering                                                                                                                               
  \setlength{\tabcolsep}{3.0pt}                                                                                                            
  \renewcommand{\arraystretch}{1.05}                                                                                                       
  \caption{Full fine-grained results on \textbf{RealDeg-Bench} (BAcc \%). The Average is computed by first averaging within each of the    
  three groups (Clean, Multi-Kind Degradation, and Compound Degradation) and then averaging the three group means. Bold indicates the best 
  Average; underline indicates the second best Average.}
  \label{tab:realdeg-full}                                                                                                                 
  \vspace{3pt}                                                                                                   

  \resizebox{0.96\columnwidth}{!}{  \begin{tabular}{>{\raggedright\arraybackslash}p{2.4cm}cccccccccccccc}
  \toprule                                                                                                                                 
  Degradation & NPR & \makecell{UnivFD} & C2P-CLIP & FatFormer & SAFE & AIDE & Effort & DRCT & Aligned & B-Free & DDA & GAPL &
  \ours{Ours-d2} & \ours{Ours-d3} \\                                                                                                       
  \midrule
  Clean & 54.05 & 50.15 & 52.89 & 57.93 & 49.62 & 63.79 & 50.70 & 74.79 & 47.46 & 84.44 & 84.77 & 86.31 & 89.31 & 87.77 \\                 
  \midrule                                                                                                                                 
  \multicolumn{15}{c}{\textit{Multi-Kind Degradation}} \\
  \midrule                                                                                                                                 
  Jpeg           & 54.10 & 50.12 & 52.82 & 56.46 & 49.53 & 63.92 & 50.71 & 74.75 & 47.40 & 84.40 & 84.84 & 86.19 & 89.35 & 87.79 \\
  Gaussian blur  & 53.73 & 47.31 & 56.50 & 60.39 & 50.82 & 50.48 & 50.31 & 64.12 & 51.37 & 76.95 & 79.09 & 81.59 & 69.95 & 84.65 \\        
  Resize         & 53.96 & 48.49 & 51.60 & 64.92 & 50.83 & 55.69 & 50.93 & 69.55 & 53.99 & 73.89 & 80.17 & 85.92 & 87.08 & 89.63 \\
  Gaussian noise & 53.56 & 46.86 & 52.65 & 46.10 & 49.37 & 57.92 & 52.91 & 73.92 & 41.64 & 81.56 & 70.53 & 75.37 & 70.37 & 82.06 \\        
  Brightness     & 56.55 & 48.38 & 52.06 & 53.39 & 49.60 & 62.53 & 51.94 & 74.65 & 47.37 & 84.36 & 85.15 & 86.64 & 89.68 & 88.63 \\        
  Contrast       & 55.94 & 48.71 & 55.45 & 49.81 & 50.31 & 58.60 & 52.76 & 73.75 & 47.76 & 83.97 & 84.51 & 83.18 & 89.53 & 87.69 \\        
  Saturation     & 51.14 & 49.68 & 52.85 & 49.77 & 50.51 & 61.71 & 50.37 & 72.25 & 46.57 & 84.54 & 85.32 & 84.75 & 87.12 & 85.30 \\        
  \midrule                                                                                                                                 
  \multicolumn{15}{c}{\textit{Compound Degradation}} \\                                                                                    
  \midrule                                                                                                                                 
  Step1 & 54.63 & 48.41 & 53.48 & 54.56 & 50.05 & 58.86 & 51.68 & 71.96 & 48.36 & 81.80 & 81.40 & 83.89 & 83.54 & 86.87 \\
  Step2 & 54.65 & 46.75 & 53.60 & 54.32 & 50.33 & 56.23 & 52.51 & 69.44 & 48.10 & 78.60 & 78.07 & 79.87 & 79.34 & 85.18 \\                 
  Step3 & 54.55 & 47.06 & 53.00 & 54.71 & 50.24 & 55.82 & 52.58 & 66.32 & 48.44 & 76.59 & 74.52 & 76.14 & 74.58 & 82.81 \\                 
  Step4 & 55.69 & 45.98 & 53.36 & 54.23 & 50.08 & 56.73 & 51.97 & 64.08 & 48.52 & 74.49 & 71.16 & 72.38 & 72.14 & 81.27 \\                 
  Step5 & 55.26 & 46.56 & 52.59 & 53.35 & 50.24 & 55.34 & 52.44 & 62.28 & 47.62 & 72.94 & 68.59 & 71.27 & 69.24 & 79.53 \\                 
  \midrule                                                                                                                                 
  \textbf{Average} & 54.38 & 48.54 & 53.17 & 55.53 & 49.98 & 59.69 & 51.45 & 71.15 & 47.90 & 80.90 & 80.30 & 82.13 & \second{82.79} &      
  \best{85.81} \\                                                                                                                          
  \bottomrule                                                                                                    
  \end{tabular}                                                                                                                            
  }                                                                                                              
  \vspace{2pt}
  \end{table}

\section{Experiments}
\label{sec:experiments}

We conduct extensive experiments to validate the effectiveness and robustness of GlobalForge on both controlled degradation and in-the-wild settings.
\subsection{Experimental Settings}

\label{sec:exp-settings}

\paragraph{Datasets.} We evaluate on two complementary suites.
\textit{(1) Robustness suite.} \textit{RealDeg-Bench} (introduced in Section~\ref{sec:realdeg-bench}) provides controllable single and multi-step degradations for fine-grained robustness diagnosis along realistic propagation chains.
\textit{(2) In-the-wild suite.} We evaluate on 8 benchmark groups: Chameleon~\cite{yan2024sanity}, SynthWildx~\cite{cozzolino2024clip}, WildRF~\cite{cavia2024real}, AIGIBench~\cite{li2025artificial}, CO-SPY~\cite{cheng2025co}, RR-Dataset~\cite{li2025bridging}, BFree-Online~\cite{guillaro2025bias}, and real-chain~\cite{liu2025beyond}. These collectively cover platform-induced degradations, evolving generators, and naturally chained social-media propagation, with no additional degradation applied at evaluation time, so the results faithfully reflect open-distribution conditions.

\paragraph{Evaluation Protocol.} We adopt Balanced Accuracy (BAcc) as the primary metric, which provides a reliable and class-balanced assessment of detection performance. We compare GlobalForge with 12 representative detectors, covering both artifact-driven methods~\cite{tan2024rethinking,ojha2023towards,liu2024forgery,li2025improving,yan2024sanity,yan2024effort,tan2025c2p} and alignment-based approaches~\cite{chen2024drct,rajan2024aligned,guillaro2025bias,chen2025dual,qin2025scaling}. Six baselines~\cite{tan2024rethinking,ojha2023towards,liu2024forgery,li2025improving,yan2024sanity,yan2024effort} are retrained from scratch on the aligned dataset~\cite{chen2025dual} to prevent detectors from latching onto format level shortcuts, while the remaining six~\cite{tan2025c2p,chen2024drct,rajan2024aligned,guillaro2025bias,chen2025dual,qin2025scaling} are evaluated with their officially released weights, as their training is closely related to their own proposed datasets or specialized objectives and cannot be cleanly transferred to a unified pipeline without distorting the method. A detailed comparison between official and retrained weights, together with the rationale for this division, is provided in Appendix~\ref{app:official-vs-ours}.

\paragraph{Implementation Details.} Following the experimental settings of DDA~\cite{chen2025dual}, we build GlobalForge on two pretrained ViT-L backbones from different families, DINOv2-L~\cite{oquab2024dinov2} and DINOv3-L~\cite{simeoni2025dinov3}, and fine-tune them with LoRA~\cite{hu2022lora}. The training data, shared with the six retrained baselines, consists of real images paired with their VAE reconstructed counterparts as fake samples. We refer to these two variants as \ours{Ours-d2} and \ours{Ours-d3} throughout the experiments. Both variants are trained on the DDA-aligned dataset~\cite{chen2025dual} described above. Full hyperparameters and training schedules are reported in Appendix~\ref{app:impl-details}.

\subsection{Controllable Degradation Evaluation on \textit{RealDeg-Bench}}
\label{sec:exp-realdeg}

We evaluate all methods on RealDeg-Bench following the protocol in Appendix~\ref{app:realdeg-bench}. The benchmark covers two regimes. \textit{Multi-Kind Degradation} applies one of the seven operators independently at multiple strengths, isolating the effect of each degradation type. \textit{Compound Degradation} applies $k$ operators in sequence ($k=1,\dots,5$), with each operator sampled from the degradation pool to simulate realistic multi-step propagation. \textit{RealDeg-Bench} itself is built on T2I-CoReBench~\cite{li2025easier}, with real images taken from its real subset and about $4{,}000$ fake images uniformly sampled per generator.

\begin{figure*}[t]
\includegraphics[width=\textwidth]{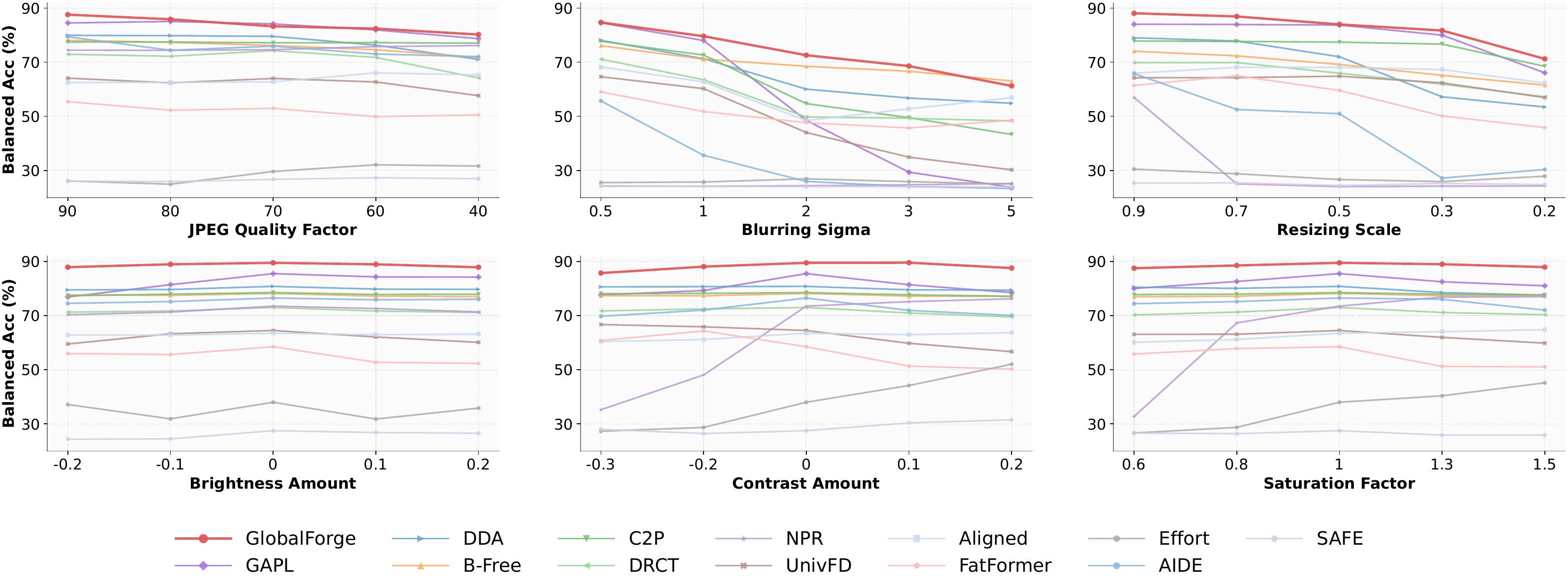}
\caption{Fine-grained strength performance curves on single degradations on \textbf{RealDeg-Bench}. The horizontal axis denotes the strength level and the vertical axis denotes B.Acc (\%).}
\label{fig:single-deg-curves}
\end{figure*}

Table~\ref{tab:realdeg-full} reports fine grained results. GlobalForge achieves the best or second best score under Clean, the vast majority of Multi-Kind Degradations, and \emph{all} five Compound Degradation levels, with an average BAcc of \textbf{85.81\%}. As shown in Figure~\ref{fig:single-deg-curves}, GlobalForge maintains a more stable trend as the strength of each single operator grows, while GAPL, DDA, and B-Free exhibit pronounced step-like drops, reflecting the fragility of local-texture cues under heavier perturbations. The advantage becomes more pronounced under Compound Degradation, where baselines suffer cliff-like drops as each additional operator destroys the local artifacts they rely on, whereas GlobalForge decays smoothly from Step1 to Step5, confirming the stability of global structural features under sequential degradation.

\subsection{In-the-Wild Comprehensive Evaluation}
\label{sec:exp-wild}

\begin{table*}[t]
\centering
\caption{Performance comparison on In-the-wild Benchmarks(BAcc \%). The Avg B.Acc is computed by first averaging within each of the 8 benchmark groups and then averaging the 8 group means. Bold indicates the best Avg B.Acc; underline indicates the second-best Avg B.Acc.}
\label{tab:wild-main}
\vspace{3pt} 
\vspace{-2mm}
\fontsize{8.2pt}{10pt}\selectfont
\setlength{\tabcolsep}{3.4pt}
\renewcommand{\arraystretch}{1.05}

\begin{adjustbox}{max width=\textwidth}
\begin{tabular}{l | c | c c c | c c c | c c | c c c c c | c | c | c | c}
\toprule
\multirow{2}{*}{\textbf{Method}}
& \multirow{2}{*}{\textbf{Cham.}}
& \multicolumn{3}{c|}{\textbf{SynthWildx}}
& \multicolumn{3}{c|}{\textbf{WildRF}}
& \multicolumn{2}{c|}{\textbf{AIGIBench}}
& \multicolumn{5}{c|}{\textbf{CO-SPY}}
& \multirow{2}{*}{\textbf{\makecell{RR\\Dataset}}}
& \multirow{2}{*}{\textbf{\makecell{BFree\\Online}}}
& \multirow{2}{*}{\textbf{\makecell{Real-\\chain}}}
& \multirow{2}{*}{\textbf{\makecell{Avg\\B.Acc}}} \\
\cmidrule(lr){3-5}
\cmidrule(lr){6-8}
\cmidrule(lr){9-10}
\cmidrule(lr){11-15}
&
& \textbf{DALLE3} & \textbf{Firefly} & \textbf{Midj.}
& \textbf{FB} & \textbf{Reddit} & \textbf{Twitter}
& \textbf{SocRF} & \textbf{ComAI}
& \textbf{Civitai} & \textbf{DALLE3} & \textbf{instavibe} & \textbf{Lexica} & \textbf{Midj.v6}
& & & & \\
\midrule
AIDE~\cite{yan2024sanity}
  & 64.48 & 68.99 & 52.84 & 84.84 & 67.50 & 66.93 & 72.30 & 60.62 & 68.69 & 90.80 & 43.55 & 13.70 & 27.90 & 94.90 &
  56.91 & 67.35 & 45.09 & 61.31 \\

  Aligned~\cite{rajan2024aligned}
  & 61.30 & 49.60 & 53.90 & 52.40 & 48.40 & 54.00 & 40.60 & 51.00 & 60.20 & 7.50 & 8.40 & 6.20 & 11.50 & 9.70 & 47.70 &
  38.10 & 49.13 & 45.02 \\

  DDA~\cite{chen2025dual}
  & 83.50 & 91.10 & 84.70 & 91.60 & 85.30 & 82.50 & 89.30 & 79.90 & 88.90 & 99.70 & 94.60 & 35.40 & 73.70 & 97.90 &
  70.30 & 81.20 & 65.81 & 80.04 \\

  B-Free~\cite{guillaro2025bias}
  & 78.30 & 96.10 & 92.30 & 95.30 & 95.60 & 85.50 & 96.70 & 84.90 & 79.70 & 98.80 & 97.40 & 5.90 & 53.60 & 98.60 & 69.50
  & 87.10 & 57.91 & 79.14 \\

  C2P-CLIP~\cite{tan2025c2p}
  & 57.60 & 49.60 & 57.90 & 49.60 & 51.90 & 67.60 & 40.40 & 58.10 & 50.40 & 0.00 & 4.50 & 0.20 & 0.20 & 23.20 & 50.00 &
  32.70 & 51.30 & 44.64 \\

  DRCT~\cite{chen2024drct}
  & 79.80 & 85.90 & 58.90 & 90.50 & 90.30 & 66.80 & 79.60 & 71.30 & 84.60 & 99.10 & 82.10 & 22.00 & 55.80 & 94.80 &
  58.20 & 77.10 & 64.43 & 73.20 \\

    NPR~\cite{tan2024rethinking}                                                                                  
    & 62.73 & 64.09 & 56.30 & 62.35 & 63.75 & 60.40 & 56.92 & 60.18 & 64.26 & 72.90 & 71.78 & 62.80 & 77.20 & 71.60 & 57.24 &              
    55.95 & 45.55 & 59.53 \\

  SAFE~\cite{li2025improving}
  & 51.80 & 53.62 & 52.96 & 53.61 & 50.31 & 53.73 & 53.08 & 52.52 & 54.81 & 99.70 & 98.13 & 95.40 & 98.70 & 98.30 &
  56.76 & 55.19 & 63.09 & 60.54 \\

  UnivFD~\cite{ojha2023towards}
  & 74.46 & 53.04 & 57.12 & 55.72 & 46.88 & 50.07 & 53.05 & 52.80 & 79.55 & 55.90 & 82.62 & 66.50 & 96.40 & 61.00 &
  49.41 & 69.68 & 62.27 & 62.47 \\

  GAPL~\cite{qin2025scaling}
  & 69.28 & 87.22 & 87.49 & 84.71 & 88.13 & 82.13 & 88.92 & 80.10 & 75.96 & 99.00 & 93.64 & 94.70 & 21.90 & 98.20 &
  78.12 & 68.40 & 63.45 & 76.45 \\

  Effort~\cite{yan2024effort}
  & 69.03 & 55.20 & 56.54 & 55.67 & 45.31 & 51.13 & 48.18 & 50.72 & 73.88 & 99.10 & 97.76 & 87.20 & 99.10 & 85.10 &
  52.61 & 59.30 & 65.36 & 63.28 \\

  FatFormer~\cite{liu2024forgery}
  & 62.10 & 54.53 & 57.48 & 61.75 & 62.81 & 61.47 & 58.85 & 52.75 & 64.08 & 95.90 & 75.14 & 67.40 & 81.90 & 76.60 &
  50.44 & 54.16 & 53.45 & 59.62 \\
\midrule

\rowcolor{blue!5}
\ours{Ours-d2}
& 88.05 & 93.78 & 85.25 & 92.20 & 86.19 & 87.73 & 95.80 & 82.68 & 93.19 & 99.60 & 99.50 & 55.70 & 93.70 & 99.30 & 74.22 & 79.99 & 65.77 & \second{83.23} \\

\rowcolor{blue!5}
\ours{Ours-d3}
& 88.03 & 92.69 & 90.18 & 92.12 & 91.88 & 94.40 & 95.90 & 87.18 & 91.55 & 99.80 & 99.25 & 65.50 & 93.20 & 98.70 & 75.93 & 84.21 & 72.86 & \best{85.93} \\

\bottomrule
\end{tabular}
\end{adjustbox}

\vspace{2pt}
\vspace{-2mm}
\end{table*}

We evaluate GlobalForge and 12 baselines on 17 in-the-wild test subsets across 8 benchmark groups~\cite{yan2024sanity,cozzolino2024clip,cavia2024real,li2025artificial,cheng2025co,li2025bridging,guillaro2025bias,liu2025beyond}, without applying any additional degradation, so the results reflect natural open-distribution conditions.

Table~\ref{tab:wild-main} shows that GlobalForge achieves the best average BAcc of \textbf{85.93\%}, outperforming the strongest baseline DDA by $5.89\%$. It ranks first on 3 of 8 groups and remains highly competitive on the others, showing consistent robustness across platforms and generators. Notably, while some baselines collapse on out-of-distribution subsets, GlobalForge maintains strong performance under diverse real world propagation conditions.

\subsection{Ablation and Mechanism Analysis}
\label{sec:ablation}

\paragraph{Component necessity.} In Table~\ref{tab:ablation-components}, baseline is the plain fine-tuned backbone without any proposed component, trained with the exact same data, backbone and other settings listed in Appendix~\ref{app:impl-details} as GlobalForge. Removing any of LIB, GSR, and DCS leads to a clear performance drop, while combining them yields complementary improvements beyond each individual component. The combination of LIB and GSR is especially effective, jointly suppressing local shortcuts and enforcing long-range reasoning.

\begin{table}[t]
\centering

\begin{minipage}[t]{0.58\linewidth}
\centering
\scriptsize
\setlength{\tabcolsep}{3pt}
\renewcommand{\arraystretch}{1.15}
\caption{Ablation of GlobalForge core components on \textbf{In-the-Wild Benchmarks} and \textbf{RealDeg-Bench} (BAcc \%). Left: single component activation; right: single component removal from the full model.}
\label{tab:ablation-components}
\vspace{2pt}
\resizebox{\linewidth}{!}{\begin{tabular}{lcc lcc}
\toprule
\multicolumn{3}{c}{\textbf{only}} & \multicolumn{3}{c}{\textbf{w/o}} \\
\cmidrule(r){1-3}\cmidrule(l){4-6}
Config. & In-the-wild & RealDeg & Config. & In-the-wild & RealDeg \\
\midrule
LIB only & 81.23 & 81.47 & w/o LIB & 83.03 & 83.80 \\
GSR only & 82.31 & 82.74 & w/o GSR & 83.12 & 83.97 \\
DCS only & 80.98 & 80.24 & w/o DCS & 83.94 & 82.61 \\
\midrule
\textbf{Full} & \textbf{85.93} & \textbf{85.81} & Baseline & 80.21 & 80.03 \\
\bottomrule
\end{tabular}}
\end{minipage}
\hfill
\begin{minipage}[t]{0.40\linewidth}
\centering
\scriptsize
\setlength{\tabcolsep}{3pt}
\renewcommand{\arraystretch}{1.15}
\caption{Alternative analysis of input-side suppression and receptive-field variants on \textbf{RealDeg-1Step} (BAcc \%). Each variant replaces one module while keeping the others intact.}
\label{tab:ablation-realdeg-extra}
\vspace{2pt}
\resizebox{\linewidth}{!}{\begin{tabular}{llc}
\toprule
Configuration & Components & BAcc \\
\midrule
Data Aug          & none         & 82.56 \\
Blur (LIB$\!\to\!$pix.) & GSR+DCS  & 81.39 \\
FFT  (LIB$\!\to\!$freq.) & GSR+DCS  & 80.12 \\
Global Attn. (GSR$\!\to\!$glob.) & LIB+DCS  & 80.73 \\
\midrule
\textbf{Full (Ours)} & \textbf{LIB+GSR+DCS} & \textbf{86.87} \\
\bottomrule
\end{tabular}}
\end{minipage}

\end{table}

\begin{figure}[H]
\centering
\includegraphics[width=\linewidth]{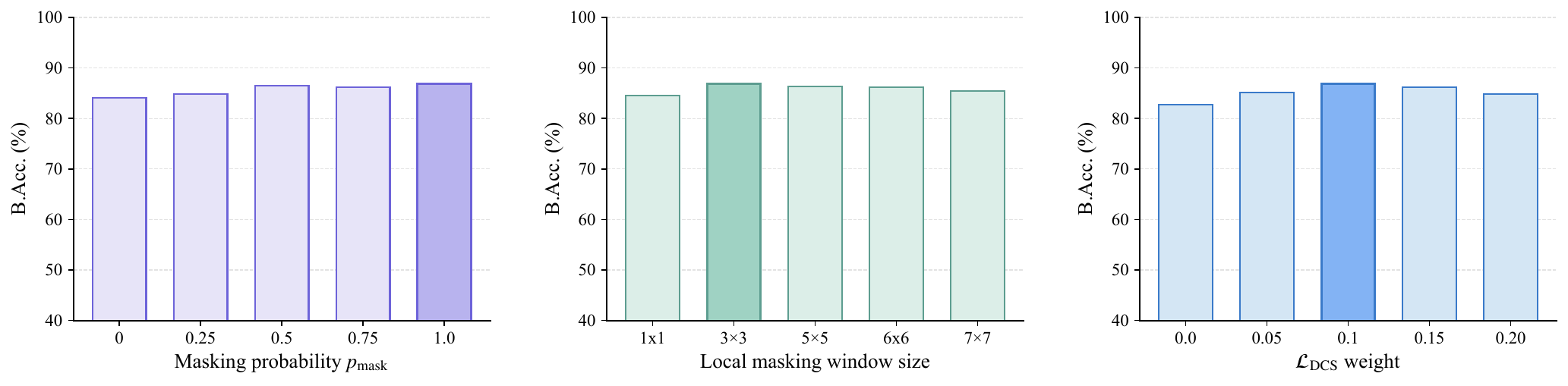}
\caption{Sensitivity analysis of our method on RealDeg-Bench under the single degradation setting, showing the impact of the masking probability $p_{\text{mask}}$, the local masking window size, and the weight of $\mathcal{L}_{\text{DCS}}$.}
\vspace{-10pt}
\label{fig:sensitivity}
\end{figure}

\paragraph{Alternative analysis.}
Table~\ref{tab:ablation-realdeg-extra} tests whether GlobalForge can be replaced by simpler alternatives. For fairness, each variant replaces only the target module while keeping all other components unchanged. Aug examines whether standard degradation augmentation alone is sufficient. Blur and FFT replace LIB with input-side suppression by pixel-domain smoothing and input-frequency low-pass filtering respectively. Consistent with the design motivation in Section~\ref{sec:method}, the results show that LIB is not merely input-side low-pass filtering. LIB operates after backbone encoding and is preferable to ordinary low-pass filtering, which can introduce new artifacts. The particularly low performance of FFT may be due to frequency artifacts coming from the operation itself, which the model may overfit as new shortcuts. Global Attention replaces GSR with unconstrained global access, showing that explicit local-window masking is needed to force long-range structural reasoning. These results indicate that GlobalForge relies jointly on degradation exposure, feature-domain local suppression, and constrained global reasoning. Detailed experimental settings are reported in Appendix~\ref{app:impl-details}.

\paragraph{Sensitivity analysis.}
\label{sec:sensitivity}
We analyze the sensitivity of our method to three hyperparameters covering both GSR and DCS: the masking probability $p_{\text{mask}}$, the local masking window size, and the weight of $\mathcal{L}_{\text{DCS}}$, evaluated under the single-step degradation setting of RealDeg-Bench (Figure~\ref{fig:sensitivity}). Accuracy peaks at $p_{\text{mask}}=1.0$ for GSR, indicating that always forcing each token to aggregate evidence from outside its local window is the most effective. For the masking window size and the weight of $\mathcal{L}_{\text{DCS}}$, a moderate setting works best. The default configuration consistently achieves the best overall trade-off.

\section{Conclusion and Discussion}
\label{sec:conclusion}

From a feature preference perspective, we identify local artifact dependency as a key cause of robustness failures in AIGI detectors. Local artifacts are highly vulnerable to realistic degradations, so detectors relying on them remain intrinsically fragile. Common degrade then restore defenses offer only limited improvement. To address this, we propose GlobalForge, which actively suppresses such fragile local cues and guides the model toward global consistency, a more stable criterion for distinguishing real from fake images. With this shift in feature preference, GlobalForge improves both robustness to degradation and generalization to unseen generative paradigms. In addition, we introduce RealDeg-Bench, a reproducible robustness benchmark inspired by common social media propagation distortions. Code, pretrained weights, and RealDeg-Bench are released at the repository linked in the abstract. Limitations and future directions are discussed in Appendix~\ref{app:limitations}.

\bibliographystyle{unsrtnat}
\bibliography{main}

@article{mirsky2021creation,
  title={The creation and detection of deepfakes: A survey},
  author={Mirsky, Yisroel and Lee, Wenke},
  journal={ACM computing surveys (CSUR)},
  volume={54},
  number={1},
  pages={1--41},
  year={2021},
  publisher={ACM New York, NY, USA}
}

@article{tolosana2020deepfakes,
  title={Deepfakes and beyond: A survey of face manipulation and fake detection},
  author={Tolosana, Ruben and Vera-Rodriguez, Ruben and Fierrez, Julian and Morales, Aythami and Ortega-Garcia, Javier},
  journal={Information fusion},
  volume={64},
  pages={131--148},
  year={2020},
  publisher={Elsevier}
}

@inproceedings{corvi2023detection,
  title={On the detection of synthetic images generated by diffusion models},
  author={Corvi, Riccardo and Cozzolino, Davide and Zingarini, Giada and Poggi, Giovanni and Nagano, Koki and Verdoliva, Luisa},
  booktitle={ICASSP 2023-2023 IEEE International Conference on Acoustics, Speech and Signal Processing (ICASSP)},
  pages={1--5},
  year={2023},
  organization={IEEE}
}

@article{pal2024semi,
  title={Semi-truths: A large-scale dataset of ai-augmented images for evaluating robustness of ai-generated image detectors},
  author={Pal, Anisha and Kruk, Julia and Phute, Mansi and Bhattaram, Manognya and Yang, Diyi and Chau, Duen Horng and Hoffman, Judy},
  journal={Advances in Neural Information Processing Systems},
  volume={37},
  pages={118025--118051},
  year={2024}
}

@inproceedings{geirhos2018imagenet,
  title={ImageNet-trained CNNs are biased towards texture; increasing shape bias improves accuracy and robustness},
  author={Geirhos, Robert and Rubisch, Patricia and Michaelis, Claudio and Bethge, Matthias and Wichmann, Felix A and Brendel, Wieland},
  booktitle={International conference on learning representations},
  year={2018}
}

@article{geirhos2020shortcut,
  title={Shortcut learning in deep neural networks},
  author={Geirhos, Robert and Jacobsen, J{\"o}rn-Henrik and Michaelis, Claudio and Zemel, Richard and Brendel, Wieland and Bethge, Matthias and Wichmann, Felix A},
  journal={Nature Machine Intelligence},
  volume={2},
  number={11},
  pages={665--673},
  year={2020},
  publisher={Nature Publishing Group UK London}
}

@inproceedings{konstantinidou2025texturecrop,
  title={Texturecrop: Enhancing synthetic image detection through texture-based cropping},
  author={Konstantinidou, Despina and Koutlis, Christos and Papadopoulos, Symeon},
  booktitle={Proceedings of the Winter Conference on Applications of Computer Vision},
  pages={1459--1468},
  year={2025}
}

@inproceedings{karageorgiou2025any,
  title={Any-resolution ai-generated image detection by spectral learning},
  author={Karageorgiou, Dimitrios and Papadopoulos, Symeon and Kompatsiaris, Ioannis and Gavves, Efstratios},
  booktitle={Proceedings of the Computer Vision and Pattern Recognition Conference},
  pages={18706--18717},
  year={2025}
}

@article{chang2023antifakeprompt,
  title={Antifakeprompt: Prompt-tuned vision-language models are fake image detectors},
  author={Chang, You-Ming and Yeh, Chen and Chiu, Wei-Chen and Yu, Ning},
  journal={arXiv preprint arXiv:2310.17419},
  year={2023}
}

@inproceedings{corvi2023intriguing,
  title={Intriguing properties of synthetic images: from generative adversarial networks to diffusion models},
  author={Corvi, Riccardo and Cozzolino, Davide and Poggi, Giovanni and Nagano, Koki and Verdoliva, Luisa},
  booktitle={Proceedings of the IEEE/CVF conference on computer vision and pattern recognition},
  pages={973--982},
  year={2023}
}

@article{oord2018representation,
  title={Representation learning with contrastive predictive coding},
  author={Oord, Aaron van den and Li, Yazhe and Vinyals, Oriol},
  journal={arXiv preprint arXiv:1807.03748},
  year={2018}
}

@inproceedings{ojha2023towards,
  title={Towards universal fake image detectors that generalize across generative models},
  author={Ojha, Utkarsh and Li, Yuheng and Lee, Yong Jae},
  booktitle={Proceedings of the IEEE/CVF conference on computer vision and pattern recognition},
  pages={24480--24489},
  year={2023}
}

@inproceedings{tan2024rethinking,
  title={Rethinking the up-sampling operations in cnn-based generative network for generalizable deepfake detection},
  author={Tan, Chuangchuang and Zhao, Yao and Wei, Shikui and Gu, Guanghua and Liu, Ping and Wei, Yunchao},
  booktitle={Proceedings of the IEEE/CVF conference on computer vision and pattern recognition},
  pages={28130--28139},
  year={2024}
}

@inproceedings{wang2023dire,
  title={Dire for diffusion-generated image detection},
  author={Wang, Zhendong and Bao, Jianmin and Zhou, Wengang and Wang, Weilun and Hu, Hezhen and Chen, Hong and Li, Houqiang},
  booktitle={Proceedings of the IEEE/CVF International Conference on Computer Vision},
  pages={22445--22455},
  year={2023}
}

@inproceedings {cozzolino2024clip,
author = { Cozzolino, Davide and Poggi, Giovanni and Corvi, Riccardo and Niesner, Matthias and Verdoliva, Luisa },
booktitle = { 2024 IEEE/CVF Conference on Computer Vision and Pattern Recognition Workshops (CVPRW) },
title = {{ Raising the Bar of AI-generated Image Detection with CLIP }},
year = {2024},
pages = {4356-4366},
doi = {10.1109/CVPRW63382.2024.00439},
url = {https://doi.ieeecomputersociety.org/10.1109/CVPRW63382.2024.00439},
publisher = {IEEE Computer Society},
address = {Los Alamitos, CA, USA},
month =Jun}

@inproceedings{frank2020leveraging,
  title={Leveraging frequency analysis for deep fake image recognition},
  author={Frank, Joel and Eisenhofer, Thorsten and Sch{\"o}nherr, Lea and Fischer, Asja and Kolossa, Dorothea and Holz, Thorsten},
  booktitle={International conference on machine learning},
  pages={3247--3258},
  year={2020},
  organization={PMLR}
}

@article{li2025easier,
  title={Easier Painting Than Thinking: Can Text-to-Image Models Set the Stage, but Not Direct the Play?},
  author={Li, Ouxiang and Wang, Yuan and Hu, Xinting and Huang, Huijuan and Chen, Rui and Ou, Jiarong and Tao, Xin and Wan, Pengfei and Qi, Xiaojuan and Feng, Fuli},
  journal={arXiv preprint arXiv:2509.03516},
  year={2025}
}

@article{oquab2024dinov2,
title={{DINO}v2: Learning Robust Visual Features without Supervision},
author={Maxime Oquab and Timoth{\'e}e Darcet and Th{\'e}o Moutakanni and Huy V. Vo and Marc Szafraniec and Vasil Khalidov and Pierre Fernandez and Daniel HAZIZA and Francisco Massa and Alaaeldin El-Nouby and Mido Assran and Nicolas Ballas and Wojciech Galuba and Russell Howes and Po-Yao Huang and Shang-Wen Li and Ishan Misra and Michael Rabbat and Vasu Sharma and Gabriel Synnaeve and Hu Xu and Herve Jegou and Julien Mairal and Patrick Labatut and Armand Joulin and Piotr Bojanowski},
journal={Transactions on Machine Learning Research},
issn={2835-8856},
year={2024},
url={https://openreview.net/forum?id=a68SUt6zFt}
}

@inproceedings{hu2022lora,
title={Lo{RA}: Low-Rank Adaptation of Large Language Models},
author={Edward J Hu and yelong shen and Phillip Wallis and Zeyuan Allen-Zhu and Yuanzhi Li and Shean Wang and Lu Wang and Weizhu Chen},
booktitle={International Conference on Learning Representations},
year={2022},
url={https://openreview.net/forum?id=nZeVKeeFYf9}
}

@inproceedings{wang2020cnn,
  title={CNN-generated images are surprisingly easy to spot... for now},
  author={Wang, Sheng-Yu and Wang, Oliver and Zhang, Richard and Owens, Andrew and Efros, Alexei A},
  booktitle={Proceedings of the IEEE/CVF conference on computer vision and pattern recognition},
  pages={8695--8704},
  year={2020}
}

@article{chen2025dual,
  title={Dual data alignment makes ai-generated image detector easier generalizable},
  author={Chen, Ruoxin and Xi, Junwei and Yan, Zhiyuan and Zhang, Ke-Yue and Wu, Shuang and Xie, Jingyi and Chen, Xu and Xu, Lei and Guan, Isabel and Yao, Taiping and others},
  journal={arXiv preprint arXiv:2505.14359},
  year={2025}
}

@inproceedings{li2025improving,
  title={Improving synthetic image detection towards generalization: An image transformation perspective},
  author={Li, Ouxiang and Cai, Jiayin and Hao, Yanbin and Jiang, Xiaolong and Hu, Yao and Feng, Fuli},
  booktitle={Proceedings of the 31st ACM SIGKDD Conference on Knowledge Discovery and Data Mining V. 1},
  pages={2405--2414},
  year={2025}
}

@inproceedings{tan2025c2p,
  title={C2p-clip: Injecting category common prompt in clip to enhance generalization in deepfake detection},
  author={Tan, Chuangchuang and Tao, Renshuai and Liu, Huan and Gu, Guanghua and Wu, Baoyuan and Zhao, Yao and Wei, Yunchao},
  booktitle={Proceedings of the AAAI Conference on Artificial Intelligence},
  volume={39},
  number={7},
  pages={7184--7192},
  year={2025}
}

@inproceedings{liu2024forgery,
  title={Forgery-aware adaptive transformer for generalizable synthetic image detection},
  author={Liu, Huan and Tan, Zichang and Tan, Chuangchuang and Wei, Yunchao and Wang, Jingdong and Zhao, Yao},
  booktitle={Proceedings of the IEEE/CVF Conference on Computer Vision and Pattern Recognition},
  pages={10770--10780},
  year={2024}
}

@article{zhu2023genimage,
  title={Genimage: A million-scale benchmark for detecting ai-generated image},
  author={Zhu, Mingjian and Chen, Hanting and Yan, Qiangyu and Huang, Xudong and Lin, Guanyu and Li, Wei and Tu, Zhijun and Hu, Hailin and Hu, Jie and Wang, Yunhe},
  journal={Advances in neural information processing systems},
  volume={36},
  pages={77771--77782},
  year={2023}
}

@article{zhong2023patchcraft,
  title={Patchcraft: Exploring texture patch for efficient ai-generated image detection},
  author={Zhong, Nan and Xu, Yiran and Li, Sheng and Qian, Zhenxing and Zhang, Xinpeng},
  journal={arXiv preprint arXiv:2311.12397},
  year={2023}
}

@article{cavia2024real,
  title={Real-time deepfake detection in the real-world},
  author={Cavia, Bar and Horwitz, Eliahu and Reiss, Tal and Hoshen, Yedid},
  journal={arXiv preprint arXiv:2406.09398},
  year={2024}
}

@inproceedings{guillaro2025bias,
  title={A bias-free training paradigm for more general ai-generated image detection},
  author={Guillaro, Fabrizio and Zingarini, Giada and Usman, Ben and Sud, Avneesh and Cozzolino, Davide and Verdoliva, Luisa},
  booktitle={Proceedings of the Computer Vision and Pattern Recognition Conference},
  pages={18685--18694},
  year={2025}
}

@ARTICLE{bammey2024synthbuster,
  author={Bammey, Quentin},
  journal={IEEE Open Journal of Signal Processing}, 
  title={Synthbuster: Towards Detection of Diffusion Model Generated Images}, 
  year={2024},
  volume={5},
  number={},
  pages={1-9},
  doi={10.1109/OJSP.2023.3337714}}

@inproceedings{chen2024drct,
  title={Drct: Diffusion reconstruction contrastive training towards universal detection of diffusion generated images},
  author={Chen, Baoying and Zeng, Jishen and Yang, Jianquan and Yang, Rui},
  booktitle={Forty-first International Conference on Machine Learning},
  year={2024}
}

@inproceedings{cheng2025co,
  title={Co-spy: Combining semantic and pixel features to detect synthetic images by ai},
  author={Cheng, Siyuan and Lyu, Lingjuan and Wang, Zhenting and Zhang, Xiangyu and Sehwag, Vikash},
  booktitle={Proceedings of the Computer Vision and Pattern Recognition Conference},
  pages={13455--13465},
  year={2025}
}

@article{ke2023df,
  title={DF-UDetector: An effective method towards robust deepfake detection via feature restoration},
  author={Ke, Jianpeng and Wang, Lina},
  journal={Neural Networks},
  volume={160},
  pages={216--226},
  year={2023},
  publisher={Elsevier}
}

@article{li2025artificial,
  title={Is artificial intelligence generated image detection a solved problem?},
  author={Li, Ziqiang and Yan, Jiazhen and He, Ziwen and Zeng, Kai and Jiang, Weiwei and Xiong, Lizhi and Fu, Zhangjie},
  journal={arXiv preprint arXiv:2505.12335},
  year={2025}
}

@inproceedings{li2025bridging,
  title={Bridging the Gap Between Ideal and Real-world Evaluation: Benchmarking AI-Generated Image Detection in Challenging Scenarios},
  author={Li, Chunxiao and Wang, Xiaoxiao and Li, Meiling and Miao, Boming and Sun, Peng and Zhang, Yunjian and Ji, Xiangyang and Zhu, Yao},
  booktitle={Proceedings of the IEEE/CVF International Conference on Computer Vision},
  pages={20379--20389},
  year={2025}
}

@inproceedings{lin2014microsoft,
  title={Microsoft coco: Common objects in context},
  author={Lin, Tsung-Yi and Maire, Michael and Belongie, Serge and Hays, James and Perona, Pietro and Ramanan, Deva and Doll{\'a}r, Piotr and Zitnick, C Lawrence},
  booktitle={European conference on computer vision},
  pages={740--755},
  year={2014},
  organization={Springer}
}

@article{liu2025beyond,
  title   = {Beyond Artifacts: Real-Centric Envelope Modeling for Reliable AI-Generated Image Detection},
  author  = {Liu, Ruiqi and Han, Yi and Zhang, Zhengbo and Yao, Liwei and Yan, Zhiyuan and Shen, Jialiang and Chen, ZhiJin and Sun, Boyi and Weng, Lubin and Dong, Jing and others},
  journal = {arXiv preprint arXiv:2512.20937},
  year    = {2025}
}

@article{qin2025scaling,
  title   = {Scaling Up AI-Generated Image Detection via Generator-Aware Prototypes},
  author  = {Qin, Ziheng and Ji, Yuheng and Tao, Renshuai and Tian, Yuxuan and Liu, Yuyang and Wang, Yipu and Zheng, Xiaolong},
  journal = {arXiv preprint arXiv:2512.12982},
  year    = {2025}
}

@article{rajan2024aligned,
  title   = {Aligned datasets improve detection of latent diffusion-generated images},
  author  = {Rajan, Anirudh Sundara and Ojha, Utkarsh and Schloesser, Jedidiah and Lee, Yong Jae},
  journal = {arXiv preprint arXiv:2410.11835},
  year    = {2024}
}

@article{simeoni2025dinov3,
  title   = {Dinov3},
  author  = {Sim{\'e}oni, Oriane and Vo, Huy V and Seitzer, Maximilian and Baldassarre, Federico and Oquab, Maxime and Jose, Cijo and Khalidov, Vasil and Szafraniec, Marc and Yi, Seungeun and Ramamonjisoa, Micha{\"e}l and others},
  journal = {arXiv preprint arXiv:2508.10104},
  year    = {2025}
}

@article{wu2025generalizable,
  title={Generalizable synthetic image detection via language-guided contrastive learning},
  author={Wu, Haiwei and Zhou, Jiantao and Zhang, Shile},
  journal={IEEE Transactions on Artificial Intelligence},
  year={2025},
  publisher={IEEE}
}

@article{yan2024effort,
  title   = {Effort: Efficient orthogonal modeling for generalizable ai-generated image detection},
  author  = {Yan, Zhiyuan and Wang, Jiangming and Wang, Zhendong and Jin, Peng and Zhang, Ke-Yue and Chen, Shen and Yao, Taiping and Ding, Shouhong and Wu, Baoyuan and Yuan, Li},
  journal = {arXiv preprint arXiv:2411.15633},
  volume  = {2},
  number  = {6},
  pages   = {7},
  year    = {2024}
}

@article{yan2024sanity,
  title={A sanity check for ai-generated image detection},
  author={Yan, Shilin and Li, Ouxiang and Cai, Jiayin and Hao, Yanbin and Jiang, Xiaolong and Hu, Yao and Xie, Weidi},
  journal={arXiv preprint arXiv:2406.19435},
  year={2024}
}

\newpage
\appendix

\clearpage
\section{RealDeg-Bench: Construction Details}
\label{app:realdeg-bench}
Section~\ref{sec:realdeg-bench} of the main text describes the design goal and degradation operator pool of RealDeg-Bench. This appendix provides additional details on the construction pipeline, including source data selection, strength sets of each operator, single-operator and compound sampling protocols, and dataset statistics.

\paragraph{Source data.}
RealDeg-Bench is constructed from a T2I-CoReBench~\citep{li2025easier}-derived image pool. For fake images, we randomly sample 3{,}978 images with a balanced distribution across generator-specific subsets. For real images, we use 3{,}375 images from the corresponding real-image split.

\paragraph{Degradation operators and strength sets.}
The seven atomic operators correspond to common distortions in real propagation chains. JPEG compression models platform re-encoding; resizing models resolution resampling; Gaussian blur models defocus and detail loss after repeated compression or reposting; Gaussian noise models capture and transmission noise; and brightness, contrast, and saturation simulate appearance shifts caused by platform-side rendering, screenshot re-sharing, or post-processing. Each operator is associated with a discrete set of strength levels covering a range from mild to severe:
\begin{itemize}\setlength\itemsep{1pt}
    \item JPEG quality factor: $Q \in \{90, 80, 70, 60, 40\}$
    \item Gaussian blur standard deviation: $\sigma \in \{0.5, 1, 2, 3, 5\}$
    \item Resize scale: $s \in \{0.9, 0.7, 0.5, 0.3, 0.2\}$
    \item Gaussian noise variance: $v \in \{0.0005, 0.001, 0.002, 0.005, 0.01\}$
    \item Brightness shift: $b \in \{-0.2, -0.1, 0.1, 0.2\}$
    \item Contrast shift: $c \in \{-0.3, -0.2, 0.1, 0.2\}$
    \item Saturation factor: $f \in \{0.6, 0.8, 1.3, 1.5\}$
\end{itemize}

\paragraph{Single-operator protocol.}
For each of the seven single-operator conditions, every image is transformed by one specified operator. For each image, the strength of that operator is independently and uniformly sampled from its discrete strength set, producing one degraded version per image. Different images within the same condition therefore receive different randomly drawn strengths.

\paragraph{Multi-step compound chain sampling.}
For compound degradation, we construct an $N$-step chain for $N \in \{1, 2, 3, 4, 5\}$. For each image, the chain is formed by sampling $N$ operators from the seven-operator pool and applying them sequentially. This design reflects realistic propagation, where an image may undergo multiple rounds of compression, resizing, color adjustment, or other platform-side processing as it traverses successive online channels. For every operator instance in the chain, its strength is independently and uniformly drawn from the corresponding strength set, so different occurrences of the same operator may apply different strengths.

\paragraph{Dataset statistics.}
RealDeg-Bench contains 13 conditions in total: clean, seven single-operator conditions, and five compound conditions ($N=1,\dots,5$). Each condition contains 3{,}375 real images and 3{,}978 fake images, yielding 7{,}353 images per condition and 95{,}589 images in total. The sampled operator sequence and the strength drawn for every operator instance are recorded and stored together with the degraded images, so all evaluations on RealDeg-Bench are deterministic and reproducible.

\clearpage

\begin{figure}[H]
\centering
\includegraphics[width=0.98\textwidth,height=0.88\textheight,keepaspectratio]{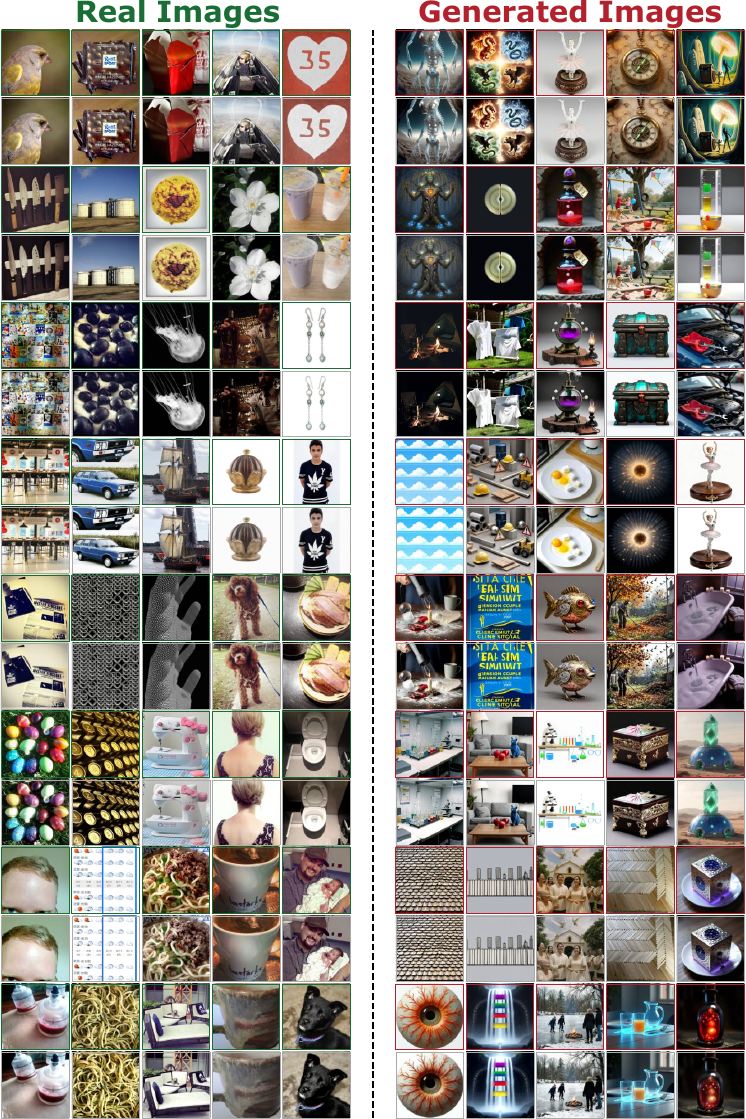}
\caption{\textbf{Example samples from \textit{RealDeg-Bench}.} Rows alternate between clean images from our dataset (odd rows) and their corresponding degraded versions (even rows), illustrating the diversity of generators, content, and degradation types covered by the benchmark.}
\label{fig:dataset-sample}
\end{figure}

\clearpage

\section{Supplementary Results on Standard Benchmarks}
\label{app:standard-benchmark}

In addition to the real-degradation and in-the-wild evaluations in the main text, we also report full numbers on common standard benchmarks as a reference for conventional detection capability. Table~\ref{tab:standard_benchmark_bacc_dacc} summarizes B.Acc and D.Acc on six public standard benchmarks: AIGCDetect~\cite{zhong2023patchcraft}, GenImage~\cite{zhu2023genimage}, UnivFD~\cite{ojha2023towards}, DRCT~\cite{chen2024drct}, Synthbuster~\cite{bammey2024synthbuster}, and EvalGEN~\cite{chen2025dual}. B.Acc is measured on clean images; D.Acc is measured after applying one randomly sampled degradation operator from our RealDeg-Bench to each benchmark. These results show that our robustness gains do not come at the cost of standard benchmark performance: GlobalForge achieves the best average D.Acc of \textbf{90.27\%} and the second-best average B.Acc of \textbf{95.45\%} across all methods.

\begin{table*}[t]
\centering
\caption{Standard benchmark results. We report Balanced Accuracy (B.Acc, \%) on clean images and Degraded Accuracy (D.Acc, \%) under one randomly sampled degradation operator from RealDeg-Bench. Bold indicates the best Avg; underline indicates the second-best Avg.}
\label{tab:standard_benchmark_bacc_dacc}
\setlength{\tabcolsep}{3.6pt}
\renewcommand{\arraystretch}{1.08}
\resizebox{\textwidth}{!}{\begin{tabular}{l cc cc cc cc cc cc cc}
\toprule
\multirow{2}{*}{\textbf{Method}}
& \multicolumn{2}{c}{\textbf{AIGCDetect}}
& \multicolumn{2}{c}{\textbf{GenImage}}
& \multicolumn{2}{c}{\textbf{UnivFD}}
& \multicolumn{2}{c}{\textbf{DRCT}}
& \multicolumn{2}{c}{\textbf{Synthbuster}}
& \multicolumn{2}{c}{\textbf{EvalGEN}}
& \multicolumn{2}{c}{\textbf{Avg}} \\
\cmidrule(lr){2-3}
\cmidrule(lr){4-5}
\cmidrule(lr){6-7}
\cmidrule(lr){8-9}
\cmidrule(lr){10-11}
\cmidrule(lr){12-13}
\cmidrule(lr){14-15}
& B.Acc & D.Acc
& B.Acc & D.Acc
& B.Acc & D.Acc
& B.Acc & D.Acc
& B.Acc & D.Acc
& B.Acc & D.Acc
& B.Acc & D.Acc \\
\midrule

GAPL~\cite{qin2025scaling}
& 97.43 & 87.82
& 96.57 & 86.84
& 90.94 & 83.89
& 96.64 & 86.85
& 95.75 & 87.09
& 97.53 & 89.36
& \best{95.81} & \second{86.97} \\

B-Free~\cite{guillaro2025bias}
& 84.66 & 79.93
& 89.63 & 82.80
& 87.77 & 81.57
& 99.25 & 95.08
& 95.74 & 92.23
& 94.60 & 89.83
& 91.94 & 86.91 \\

DDA~\cite{chen2025dual}
& 81.52 & 77.48
& 88.91 & 81.96
& 69.38 & 71.56
& 97.03 & 88.95
& 95.38 & 88.24
& 95.41 & 85.74
& 87.94 & 82.32 \\

Effort~\cite{yan2024effort}
& 71.22 & 65.42
& 86.29 & 74.87
& 55.07 & 58.06
& 89.09 & 74.87
& 88.93 & 76.09
& 97.08 & 78.12
& 81.28 & 71.24 \\

DRCT~\cite{chen2024drct}
& 67.06 & 60.74
& 78.75 & 68.71
& 66.77 & 62.62
& 97.01 & 79.69
& 80.99 & 72.82
& 81.91 & 78.60
& 78.75 & 70.53 \\

AIDE~\cite{yan2024sanity}
& 84.00 & 70.01
& 88.56 & 76.75
& 77.85 & 64.39
& 59.24 & 63.09
& 75.40 & 70.88
& 59.54 & 70.28
& 74.10 & 69.23 \\

SAFE~\cite{li2025improving}
& 48.36 & 63.62
& 47.74 & 74.37
& 49.69 & 57.43
& 50.31 & 68.78
& 51.56 & 75.65
& 50.15 & 69.18
& 49.64 & 68.17 \\

UnivFD~\cite{ojha2023towards}
& 56.46 & 69.81
& 62.51 & 64.20
& 48.28 & 71.89
& 69.50 & 61.03
& 65.74 & 65.61
& 73.78 & 60.14
& 62.71 & 65.45 \\

FatFormer~\cite{liu2024forgery}
& 82.13 & 62.05
& 71.50 & 63.86
& 86.71 & 60.89
& 53.88 & 62.13
& 69.08 & 70.18
& 56.59 & 66.07
& 69.98 & 64.20 \\

C2P-CLIP~\cite{tan2025c2p}
& 79.95 & 68.33
& 71.40 & 60.42
& 87.30 & 74.38
& 54.58 & 56.17
& 69.42 & 61.24
& 57.72 & 63.53
& 70.06 & 64.01 \\

Aligned~\cite{rajan2024aligned}
& 55.70 & 53.83
& 57.50 & 54.93
& 60.48 & 58.07
& 54.87 & 56.82
& 54.54 & 54.24
& 65.76 & 58.35
& 58.14 & 56.04 \\

  NPR~\cite{tan2024rethinking}                                                                         
  & 64.79 & 58.68                                                                                      
  & 68.49 & 60.94                                                                                      
  & 51.99 & 51.89                                                                                      
  & 69.15 & 60.45                                                                                      
  & 73.74 & 64.18                                                                                    
  & 68.23 & 58.70                                                                                      
  & 66.07 & 59.14 \\

\midrule

\rowcolor{blue!5}
\ours{GlobalForge}
& 93.30 & 87.96
& 96.76 & 90.64
& 86.90 & 85.13
& 99.22 & 93.27
& 99.41 & 95.58
& 97.14 & 89.04
& \second{95.45} & \best{90.27} \\

\bottomrule
\end{tabular}}
\end{table*}

\begin{table*}[h]
  \centering
  \caption{Comparison of \textbf{Official Weights} and \textbf{Our Retrained Weights} on in-the-wild benchmarks. We report Balanced Accuracy (B.Acc, \%) on each dataset group and the equal-weighted average over the 8 groups. \textbf{Bold} marks the better weights in Average; all \texttt{Ours} rows are trained from scratch on the same MSCOCO~\cite{lin2014microsoft} + AIGI split as our model.}
  \label{tab:official_vs_ours_wild}
  \setlength{\tabcolsep}{4pt}
  \renewcommand{\arraystretch}{1.1}
  \resizebox{\textwidth}{!}{  \begin{tabular}{l c c c c c c c c c c}
  \toprule
  \textbf{Method} & \textbf{Weights} & \textbf{Chameleon} & \textbf{SynthWildx} & \textbf{WildRF} & \textbf{AIGIBench} & \textbf{CO-SPY} & \textbf{RR-Dataset} & \textbf{BFree-Online} & \textbf{Real-Chain} & \textbf{Average} \\
  \midrule
    \multirow{2}{*}{AIDE~\cite{yan2024sanity}}                                                                                                   
     & Official & 59.35 & 57.65 & 62.49 & 58.69 & 22.23 & 52.47 & 51.76 & 40.82 & 50.68 \\                                                       
     & Ours     & 64.48 & 68.89 & 68.91 & 64.66 & 54.17 & 56.91 & 67.35 & 45.09 & \textbf{61.31} \\                                              
    \midrule                                                                                                                                     
    \multirow{2}{*}{SAFE~\cite{li2025improving}}                                                                                                 
     & Official & 52.50 & 49.93 & 51.26 & 50.40 &  0.83 & 49.81 & 50.00 & 36.45 & 42.65 \\                                                       
     & Ours     & 51.80 & 53.40 & 52.37 & 53.67 & 98.05 & 56.76 & 55.19 & 63.09 & \textbf{60.54} \\                                              
    \midrule                                                                                                                                     
    \multirow{2}{*}{NPR~\cite{tan2024rethinking}}                                                                                                
     & Official & 61.26 & 54.58 & 64.22 & 63.25 & 78.79 & 56.99 & 66.06 & 49.55 & \textbf{61.84} \\                                              
     & Ours     & 62.73 & 60.91 & 60.36 & 62.22 & 71.26 & 57.24 & 55.95 & 45.55 & 59.53 \\                                                       
    \midrule                                                                                                                                     
    \multirow{2}{*}{Effort~\cite{yan2024effort}}                                                                                                 
     & Official & 52.41 & 49.86 & 63.13 & 57.40 & 47.07 & 49.59 & 53.49 & 46.18 & 52.39 \\                                                       
     & Ours     & 69.03 & 55.80 & 48.21 & 62.30 & 93.65 & 52.61 & 59.30 & 65.36 & \textbf{63.28} \\                                              
    \midrule                                                                                                                                     
    \multirow{2}{*}{FatFormer~\cite{liu2024forgery}}                                                                                             
     & Official & 50.88 & 49.82 & 49.95 & 50.23 & 96.14 & 50.58 & 49.85 & 50.57 & 56.00 \\                                             
     & Ours     & 62.10 & 57.92 & 61.04 & 58.42 & 79.39 & 50.44 & 54.16 & 53.45 & \textbf{59.62} \\               
    \midrule                                                                                                                                     
    \multirow{2}{*}{UnivFD~\cite{ojha2023towards}}                                                                                               
     & Official & 50.18 & 51.12 & 54.98 & 52.86 &  9.10 & 46.21 & 48.81 & 35.82 & 43.64 \\
     & Ours     & 74.46 & 55.29 & 50.00 & 66.17 & 72.48 & 49.41 & 69.68 & 62.27 & \textbf{62.47} \\                                              
    \bottomrule
  \end{tabular}}
\end{table*}
\section{Comparison between Official Weights and Our Retrained Weights}
\label{app:official-vs-ours}

In the in-the-wild evaluation of Section~\ref{sec:exp-wild}, the 12 baselines are evaluated under two settings depending on whether their training pipelines can be cleanly transferred to a unified protocol.

For C2P-CLIP~\cite{tan2025c2p}, DRCT~\cite{chen2024drct}, Aligned~\cite{rajan2024aligned}, B-Free~\cite{guillaro2025bias}, DDA~\cite{chen2025dual}, and GAPL~\cite{qin2025scaling}, we follow standard community practice and use their officially released weights. DRCT, Aligned, B-Free, and DDA are tightly coupled with their own proposed alignment datasets, and the data curation and alignment procedure is an integral part of the training pipeline that cannot be cleanly decoupled from the detector itself. C2P-CLIP and GAPL adopt specialized training paradigms (image text contrastive learning with category common prompts and prototype mapping over canonical generators, respectively) that are incompatible with a unified binary classification pipeline. Retraining any of them on a different data source or under a different objective would fundamentally alter the approach being evaluated, so we adopt the released checkpoints to faithfully represent each method's original design.

The remaining six baselines, NPR~\cite{tan2024rethinking}, UnivFD~\cite{ojha2023towards}, FatFormer~\cite{liu2024forgery}, SAFE~\cite{li2025improving}, AIDE~\cite{yan2024sanity}, and Effort~\cite{yan2024effort}, are retrained from scratch following the experimental settings of DDA~\cite{chen2025dual}, sharing the same training data, splits, and evaluation pipeline as GlobalForge. Aside from minimal modifications required to plug into the unified pipeline, we do not alter the core code of any baseline.

Across both the official-weights and retrained-weights settings, we retain each baseline's original backbone rather than transplanting every method onto a unified backbone, because most of these detectors are designed around backbone-specific properties, and forcing them onto a uniform feature extractor would invalidate those design choices and no longer reflect the method as proposed.

As shown in Table~\ref{tab:official_vs_ours_wild}, our retrained weights outperform the officially released weights on the average of the 8 in-the-wild benchmark groups for most methods, confirming that the gains come from better data quality and training uniformity rather than any unfair advantage.

\begin{table*}[t]
\centering
\small
\setlength{\tabcolsep}{6pt}
\renewcommand{\arraystretch}{1.15}
\newcolumntype{Y}{>{\hsize=\dimexpr 2\hsize+2\tabcolsep\relax\raggedright\arraybackslash}X}
\caption{  Complete training setup of the two model variants.
  Fields marked with $\dagger$ were absent from the original table and are
  derived directly from the source code.}
\label{tab:train_setup_full}
\begin{tabularx}{\textwidth}{@{}l X X@{}}
\toprule
\textbf{Field} & \textbf{ours-d2} & \textbf{ours-d3} \\
\midrule
\multicolumn{3}{@{}l}{\textit{Backbone \& LoRA adaptation}} \\
\midrule
Backbone (frozen)              & DINOv2-L (\texttt{dinov2-large}) & DINOv3-L (\texttt{dinov3-large}) \\
Input / patch size$^\dagger$   & $224{\times}224$, patch $14$     & $224{\times}224$, patch $16$ \\
LoRA$^\dagger$                 & \multicolumn{2}{Y}{$r{=}16,\ \alpha{=}32$, dropout $0$, on Q/K/V of all 24 layers} \\
\midrule
\multicolumn{3}{@{}l}{\textit{Optimisation}} \\
\midrule
Optimizer$^\dagger$            & \multicolumn{2}{Y}{AdamW ($\beta_1{=}0.9,\ \beta_2{=}0.999,\ \epsilon{=}10^{-8}$), weight decay $0.01$} \\
LR (peak / min)                & $4{\times}10^{-4}$ / $2{\times}10^{-5}$ & $5{\times}10^{-5}$ / $1{\times}10^{-6}$ \\
LR schedule$^\dagger$          & \multicolumn{2}{Y}{Half-cycle cosine decay, 1 warm-up epoch, fp32, no grad clipping} \\
\midrule
\multicolumn{3}{@{}l}{\textit{Schedule, batch \& module hyperparameters}} \\
\midrule
GPUs                           & 1$\times$A100-80\,GB & 4$\times$A100-80\,GB \\
Epochs (scheduled / logged)    & 10 / 8               & 8 / 6 \\
Batch / GPU, accum, effective  & 128, 16, 2048        & 128, 4, 2048 \\
Samples per epoch$^\dagger$    & \multicolumn{2}{Y}{80\,000 (40k real + 40k fake, capped from 118{,}287/class); seed $0$, single run} \\
LIB$^\dagger$                  & \multicolumn{2}{Y}{$k{=}3$, $\sigma{=}1.0$; gate $\beta$ as a single learnable scalar} \\
GSR$^\dagger$                  & \multicolumn{2}{Y}{$w_{\mathrm{gsr}}{=}3$ (Chebyshev distance threshold)} \\
DCS loss$^\dagger$             & \multicolumn{2}{Y}{$\lambda_{\mathrm{dcs}}{=}0.01$,\ \ $\tau{=}0.07$} \\
DCS augmentation$^\dagger$     & \multicolumn{2}{Y}{JPEG $Q{\in}(20,80)$;\ \ blur $\sigma{\in}(0.5,1.5)$ with $7{\times}7$ kernel;\ \ color jitter $0.4$ / hue $0.06$} \\
\bottomrule
\end{tabularx}

\vspace{2pt}
\footnotesize
\textbf{Software / hardware.}~All experiments run on NVIDIA A100 80\,GB PCIe, Python~3.10.18,
PyTorch~2.8.0\,+\,cu126 (CUDA~12.6).
\end{table*}
\section{Implementation Details}
\label{app:impl-details}

For reproducibility, this section consolidates the complete training and inference setup of GlobalForge. GlobalForge follows a ``\textbf{pretrained vision backbone + lightweight structural modules + joint training objectives}'' paradigm. The input image is patchified and fed into the pretrained Transformer encoder; LIB and GSR are inserted sequentially after the backbone output; and the classification head produces the real/fake decision. Unlike many methods that rely on additional restoration branches, frequency-domain branches, or test-time ensembling, GlobalForge's main inference path remains a single forward pass, and the additional modules all operate in feature space. Hence, the key training details lie not in a complex data flow, but in the stable joint optimization of local suppression, global reasoning, and the degradation-invariance constraint.

\paragraph{Compute resources.}
All experiments were run on NVIDIA A100 80GB PCIe GPUs. The main GlobalForge variants used
1 A100 GPU for ours-d2 and 4 A100 GPUs for ours-d3. Training ours-d2 took approximately 9 GPU-hours,
and training ours-d3 took approximately 24 GPU-hours.

\paragraph{Construction of degraded views.} The degradation-aware contrastive structural loss $\mathcal{L}_{\text{DCS}}$ requires an online-constructed degraded view for each training sample. Following Section~\ref{sec:method-optim} in the main text, this view is constructed by random JPEG compression, random Gaussian blur, and random color perturbation. The goal is to create sufficiently diverse appearance perturbations for the same semantic content during training, so that the model is forced to learn representations that are stable across degradations. The image representation $\mathbf{z}_i$ is obtained by average-pooling the patch-token outputs from the final backbone layer.

\paragraph{Alternative-analysis variants.}
To support the alternative analysis in Table~\ref{tab:ablation-realdeg-extra}, we train four additional variants following the same implementation pipeline as the main model, with only the tested mechanism changed. \textit{Aug} removes LIB, GSR, and $\mathcal{L}_{\text{DCS}}$, and trains the classifier with standard random degradation augmentation only. \textit{Blur} replaces the feature-domain LIB with input Gaussian smoothing, applying a $3{\times}3$ Gaussian blur with $\sigma=1.0$ to the image input during both training and evaluation. \textit{FFT} replaces LIB with input frequency low-pass filtering, using a cutoff ratio of $0.25$ during both training and evaluation. \textit{Global Attention} keeps the attention module global but disables the GSR, thereby testing an unconstrained global receptive-field alternative. All variants are evaluated on the one-step compound split of RealDeg-Bench, denoted as RealDeg-1Step in Table~\ref{tab:ablation-realdeg-extra}.

Table~\ref{tab:train_setup_full} records the key hyperparameters actually used by GlobalForge in the main experiments, which directly reproduce the experimental setup and are useful for comparisons with other methods.

\paragraph{Dataset and asset licenses.}
All datasets and pretrained models are used in compliance with their respective licenses and solely for non-commercial academic research.
DINOv2~\cite{oquab2024dinov2} is used under Apache~2.0;
DINOv3~\cite{simeoni2025dinov3} under Meta's non-commercial research license;
GenImage~\cite{zhu2023genimage} and Synthbuster~\cite{bammey2024synthbuster} under CC~BY-NC-SA~4.0;
DRCT~\cite{chen2024drct}, EvalGEN~\cite{chen2025dual}, and T2I-CoReBench~\cite{li2025easier} under Apache~2.0;
LoRA~\cite{hu2022lora} under MIT.
AIGCDetect~\cite{zhong2023patchcraft} and UnivFD~\cite{ojha2023towards} do not publish explicit licenses and are accessed through their official repositories for research use only.
\section{Supplementary Visual Examples}
\label{app:qualitative}

Beyond Figure~\ref{fig:attention-collapse} in the main text, we provide two additional qualitative views: (i)~typical success and failure cases of our detector, and (ii)~side-by-side attention heatmaps on clean and degraded inputs for several detectors. Both views are interpreted under the same lens used throughout the paper: GlobalForge is designed to base its decision on long-range structural evidence rather than on any local artifact signature, and these visualizations are meant to make that design choice empirically visible.

\paragraph{Qualitative analysis of success and failure cases.}
\begin{figure*}[t]
  \centering
  \includegraphics[width=\linewidth]{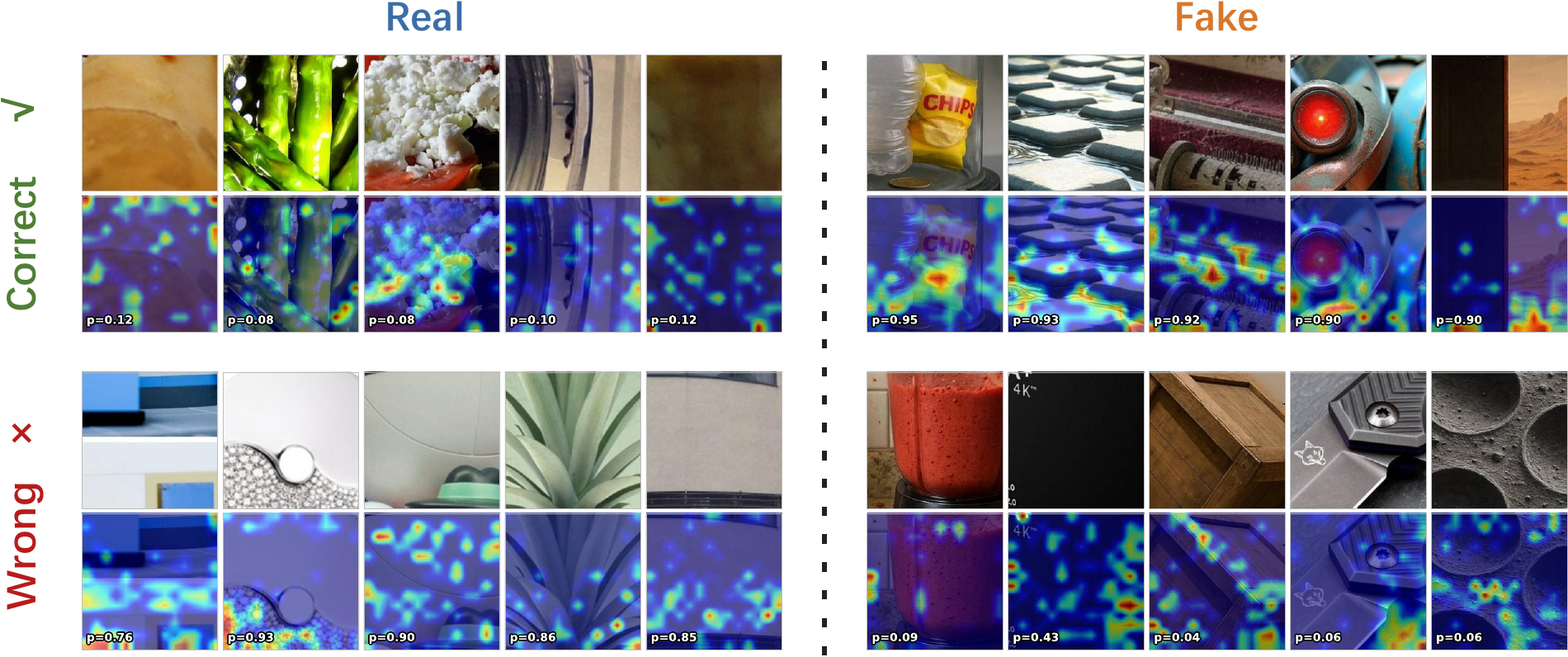}
  \caption{\textbf{Qualitative success and failure cases of GlobalForge.}
  The figure uses a $2{\times}2$ block layout (correct vs.\ wrong $\times$ real vs.\ fake). Each block contains five columns; the input image is on top and the Grad-CAM heatmap (annotated with the predicted fake probability $p$) is on the bottom. The four blocks are: correctly classified real images (top-left, $p\!\le\!0.12$), correctly classified fake images (top-right, $p\!\ge\!0.90$), real images misclassified as fake (bottom-left, $p\!>\!0.5$), and fake images misclassified as real (bottom-right, low-to-borderline $p$).}
  \label{fig:aigi_good_bad_case}
\end{figure*}
Fig.~\ref{fig:aigi_good_bad_case} shows representative success and failure cases together with the model's Grad-CAM attention. The highlighted regions mark tokens whose contribution to the final score is large; under the GSR design (Section~\ref{sec:method-gsr}), each such token has already aggregated information from distant patches outside its own local window, so the highlights should be read as anchors of a long-range structural judgement rather than as isolated local cues. In the successful cases (top row), these anchors are spread across multiple regions of the image instead of clustering on a single local patch, and the predicted probabilities are confident on both sides ($p\!\le\!0.12$ for real and $p\!\ge\!0.90$ for fake). The failure cases (bottom row) sit close to the decision boundary and share a common pattern: the image offers little exploitable long-range structure. Real images with smooth or low-texture content (e.g.\ flat color regions, close-ups of repetitive patterns) provide too few cross-region cues, so a small number of accidental responses is enough to push the score over the boundary and the image ends up labelled as fake. Fake images dominated by natural-looking objects or strong photographic priors (e.g.\ realistic lighting, plausible material reflections) instead present a globally coherent appearance whose long-range structure does not differ visibly from a real photograph, and the detector misses them. The remaining errors therefore concentrate on samples where the global-structure signal is weak or already close to that of natural images, which is consistent with our design principle of relying on global structural coherence rather than on any specific local artifact signature.

\paragraph{Stability of spatial focus under degradation.}
\begin{figure*}[t]
      \centering
      \includegraphics[width=\linewidth]{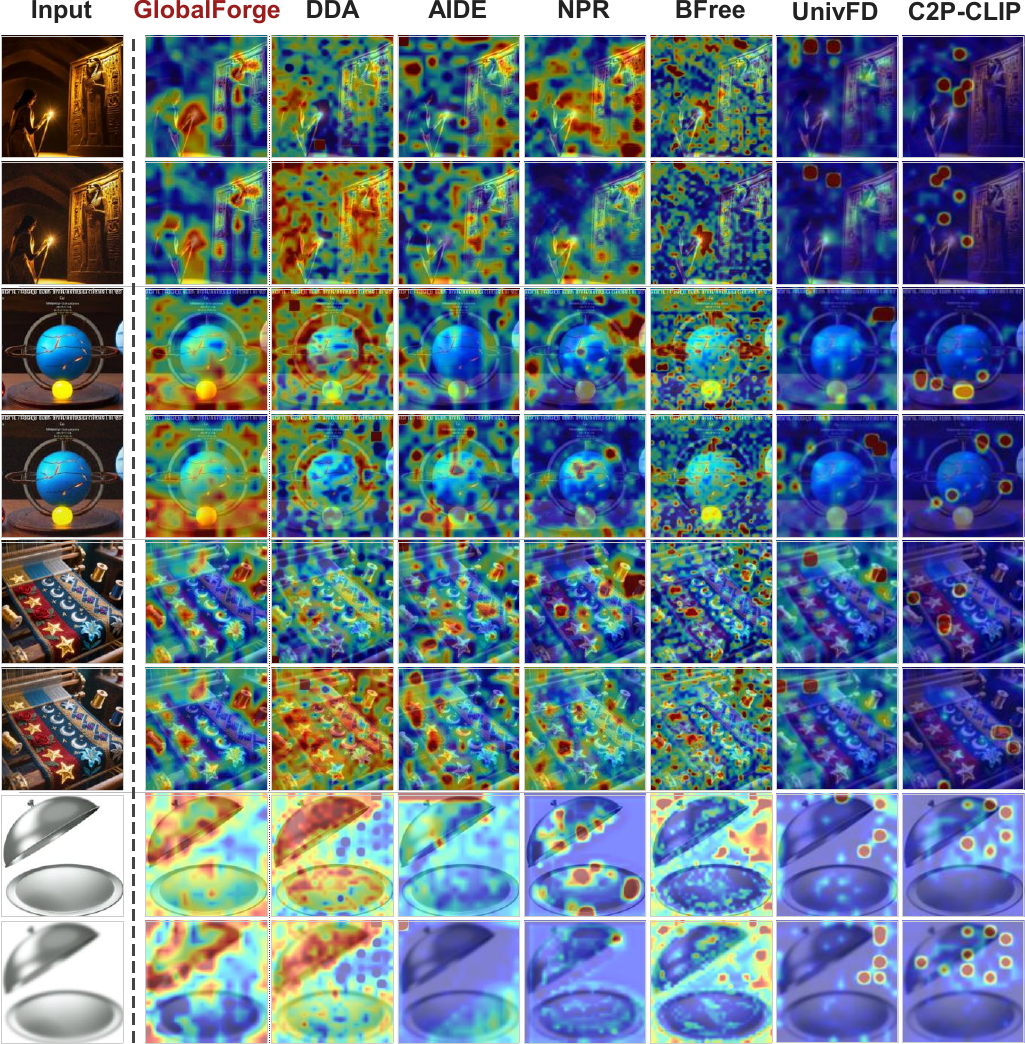}
      \caption{\textbf{Attention stability under degradation: GlobalForge versus six representative baselines.}
      Each pair of consecutive rows is one image, shown clean (top) and degraded (bottom). Columns from left to right are: the input image, GlobalForge (ours), and six competing detectors (DDA, AIDE, NPR, B-Free, UnivFD, C2P-CLIP). For each method, the two heatmaps in a pair use the same normalization range, so a visible change in the highlighted regions between the two rows is a direct sign that the method's spatial focus is unstable.}
      \label{fig:aigi_contrast_heatmap}
  \end{figure*}
Figure~\ref{fig:aigi_contrast_heatmap} compares the spatial responses of GlobalForge with six representative baselines on four image pairs, where each pair is a clean image and a degraded counterpart obtained from a single perturbation. Within a pair, the two heatmaps share a normalization range, so visual differences read as direct evidence about the stability of the method's focus. Across all four pairs, the baselines change a lot after degradation: their attended regions drift to unrelated areas, become diffuse, or collapse onto a few isolated points that no longer correspond to the regions used on the clean image. This is the visual signature of detectors whose decision is anchored on local artifacts: once a perturbation alters the local statistics, the spatial focus has nothing stable to hold on to. Our method instead keeps its attention on essentially the same regions before and after degradation, with only minor changes in intensity. We attribute this to the GSR design (Section~\ref{sec:method-gsr}), which forces every token to aggregate evidence from distant patches outside its local window: the resulting decision is supported by a long-range structural pattern that survives a single local perturbation, and the contrastive structural loss further pulls clean and degraded representations together so that the global structural cues stay aligned. This matches the quantitative results in Section~\ref{sec:exp-realdeg}: as the input statistics shift, baseline detectors lose their local discriminative cues, while GlobalForge still anchors its predictions on the global structural pattern, which is the underlying reason it degrades more gracefully.

\section{Broader Impact}
\label{app:broader-impact}

The development of robust AI-generated image (AIGI) detection methods has important societal, ethical, and technical implications. Our work contributes to mitigating the risks of synthetic visual content, including misinformation, impersonation, and non-consensual deepfakes, by enabling more reliable verification of image authenticity under realistic online propagation. Images shared on social platforms are often compressed, resized, blurred, and re-shared before inspection; by explicitly targeting these degraded in-the-wild conditions, GlobalForge provides a practical framework for detecting AI-generated images beyond clean benchmark settings. This can strengthen trust in digital media, support content moderation efforts, and assist forensic or journalistic investigations involving visual evidence.

This research also aligns with broader efforts to establish trustworthy multimedia ecosystems. By shifting detection from fragile local generator artifacts toward degradation-stable global structural evidence, GlobalForge advances a robustness-oriented detection mechanism that is better matched to real-world deployment conditions. The proposed RealDeg-Bench further supports this goal by enabling fine-grained evaluation of detector robustness under common propagation operations and compound degradation chains. We hope these contributions can facilitate more reliable auditing of AI-generated visual content while promoting public awareness of synthetic media risks.

At the same time, AIGI detection should be used with care. No detector is fully reliable, and incorrect predictions may lead to false accusations, overlooked synthetic content, or misplaced confidence in automated decisions. Therefore, we recommend treating detector outputs as supporting evidence rather than definitive judgments, especially in high-stakes settings such as journalism, legal analysis, or content moderation. Human review, provenance metadata, and contextual information should be considered alongside model predictions.

We encourage interdisciplinary collaboration among researchers, platform operators, journalists, forensic analysts, ethicists, and policymakers to ensure that AIGI detection technologies serve as safeguards for digital media authenticity and public trust.
\section{Limitations and Future Work}
\label{app:limitations}

While GlobalForge demonstrates strong robustness across controlled degradation settings and in-the-wild AIGI detection benchmarks, several limitations and opportunities for future work remain.

\paragraph{Limitations.}
First, although RealDeg-Bench covers a broad set of common propagation degradations, real-world image circulation is more diverse and continuously evolving. Platform-side compression policies, editing tools, screenshot behaviors, and re-sharing pipelines may change over time, which can introduce degradation patterns beyond the current benchmark. Second, GlobalForge is designed around the assumption that global structural cues are more stable than local artifacts. This assumption is well supported by our experiments, but future generative models may reduce such structural inconsistencies as synthesis quality improves. Third, while LIB and GSR use lightweight designs, their key hyperparameters, such as the local suppression strength and the masking window, are still manually specified. More adaptive designs may further improve robustness across datasets and resolutions.

\paragraph{Future Directions.}
Future work could extend RealDeg-Bench with dynamic re-calibration for newly emerging platform pipelines and more diverse user-side editing operations. Another promising direction is to study whether global structural cues remain reliable for more advanced generators, localized image editing, and multimodal generation scenarios. In addition, adaptive local suppression and structure-aware reasoning modules could be explored to reduce manual design choices while preserving the robustness benefits of GlobalForge. Finally, extending the framework from image-level AIGI detection to video-generated content detection may further test whether degradation-stable structural reasoning can generalize beyond static images.

\end{document}